\pdfoutput=1
\documentclass[11pt]{article}

\usepackage{EACL2023}

\usepackage{times}
\usepackage{latexsym}

\usepackage[T1]{fontenc}
\usepackage[utf8]{inputenc}

\usepackage{microtype}

\usepackage{inconsolata}

\usepackage{hyperref}
\usepackage{url}

\usepackage[pdftex]{graphicx}
\usepackage{array}
\usepackage{multirow}
\usepackage{amsmath}
\usepackage{soul}
\usepackage{amssymb}
\usepackage{booktabs}
\usepackage{threeparttable}
\usepackage[toc,page]{appendix}
\usepackage{float}
\usepackage{makecell}
\usepackage{listings}
\usepackage{comment}
\usepackage{arydshln}
\usepackage{makecell}
\usepackage{enumitem}

\newcommand{\dd}[1]{{#1}}

\newcommand{\textplus}{\raisebox{.4\height}{\scalebox{.6}{+}}}

\title{Embedding Recycling for Language Models}

\author{
  \textbf{Jon Saad-Falcon\textsuperscript{1}\quad } \textbf{Amanpreet Singh\textsuperscript{1}\quad } \textbf{Luca Soldaini\textsuperscript{1}\quad} \\ \textbf{Mike D'Arcy\textsuperscript{2}\quad } \textbf{Arman Cohan\textsuperscript{1,3}\quad } \textbf{Doug Downey\textsuperscript{1,2}}\vspace{6pt}\\
 \textsuperscript{1}Allen Institute for Artificial Intelligence (AI2) \\
 \textsuperscript{2} Northwestern University \\
 \textsuperscript{3} Yale University \\
  \texttt{\{jons, amanpreets, lucas, armanc, dougd\}@allenai.org},\\ \texttt{m.m.darcy@u.northwestern.edu}}

\begin{document}

\maketitle

\begin{abstract}

Real-world applications of neural language models often involve running many different models over the same corpus.  \dd{The resulting high computational cost} has led to interest in techniques that can reuse the contextualized embeddings produced in previous runs to speed training and inference of future ones.  We refer to this approach as {\em embedding recycling (ER).} While multiple ER techniques have been proposed, their practical effectiveness is still unknown because existing evaluations consider very few models and do not adequately account for overhead costs.  We perform an extensive evaluation of ER across eight different models (17 to 900 million parameters) and fourteen tasks in English.  We show how a simple ER technique that caches activations from an intermediate layer of a pretrained model, and learns task-specific adapters on the later layers, is broadly effective.  For the best-performing baseline in our experiments (DeBERTa-v2 XL), adding a precomputed cache results in a >90\% speedup during training and 87-91\% speedup for inference, with negligible impact on accuracy.  Our analysis reveals important areas of future work, and we release code and documentation for our experiments at \url{https://github.com/allenai/embeddingrecycling}. %, including new methods for recycling activations across heterogeneous models and reducing long-term storage cost.

\end{abstract}

\section{Introduction}

Large pretrained language models form the foundation of modern NLP, and continue to push the state-of-the-art on a wide range of natural language processing tasks \citep{devlin2018bert,liu2019roberta,bommasani2021opportunities}.  Larger models tend to offer superior accuracies \citep{kaplan2020scaling}, but also \dd{entail} higher computational costs.  The steep computational cost  associated with large neural language models slows down experimentation, increases financial barriers to the technology, and contributes to global climate change \citep{strubell2019energy, dodge2022measuring}.

\dd{Our work studies how to reduce computational cost for workloads in which many distinct models are run over the same text.}  For example, a scholarly search tool that helps users find and understand relevant literature may run separate models for entity recognition, topic classification, relation extraction, summarization, question answering, and so on over a large corpus of papers.  New and improved models for the tasks are developed frequently, necessitating additional runs.  \dd{The need for repeated model runs has also been noted for other applications in previous work, including} news applications \cite{du2020general} and virtual assistants \cite{Wei2022AFM}.  \dd{Further, r}epeated runs also occur very frequently during model development, when exploring model variants or executing multiple training epochs.

%Recently, researchers have introduced ways to reduce this computational cost by re-using activations from previous runs to speed up the current one \cite{du2020general,Wei2022AFM}.  In many important corpora, substantial portions of the text remain fixed over time (e.g. scientific papers, Wikipedia articles, StackExchange posts, financial reports, legal records, etc.), and many models are run multiple times over the text.  For example, a scholarly search tool that helps users find and understand relevant literature needs to perform entity recognition, topic classification, relation extraction, summarization, question answering, and so on over a large corpus of past papers.  New and improved models for each of the tasks are developed frequently, necessitating additional inference passes over the corpus.  A similar need to run many models on fixed text has also been noted for news applications \cite{du2020general} and virtual assistants \cite{Wei2022AFM}.  \dd{Further, r}epeated runs also occur very frequently during model development, when exploring model variants or executing multiple training epochs. %Further, individual practitioners often repeatedly train the same large model on the same text e.g. hyperparameter search \arman{This is an interesting use case. If we do hyperparam tuning under ER setting and find optimal ones, do we know if the same hyperparams would be optimal for the original model?}.

%Any one run of a pretrained language model produces a contextualized embedding of the text.
\dd{Recent work has introduced ways to reduce computational cost in such settings by re-using model activations from one task to speed up other ones \cite{du2020general,Wei2022AFM}. A pretrained language model's internal activations form a contextualized embedding, which reflects syntactic and semantic knowledge about the input text \citep{goldberg2019assessing,wiedemann2019does,rogers-etal-2020-primer} which can be useful across a variety of downstream tasks.}  %The approach we study is based on the intuition that after this contextualized embedding has been computed once for a text, it can be reused for future tasks.
We define {\em embedding recycling} (ER) as the technique of caching certain activations from a previous model run, and re-using them to improve the efficiency of future training and inference.  Recycling imposes a small computation time cost the first time a model processes a text, in order to compute and populate the cache. Thereafter, all subsequent runs on the text can use the precomputed cache, improving efficiency.  %This is significant because the number of such subsequent runs can be large in some real-world applications, including when re-training models many times during development, executing many epochs in a training run, or performing inference with many different models over the same corpus \cite{du2020general,Wei2022AFM}.

While previous work has shown the promise of ER approaches, the existing evaluations are limited.  For example, \citet{du2020general} and \citet{Wei2022AFM} \dd{each evaluate ER for only one or two base models.  Likewise, for ER techniques that cache activations on persistent storage, the storage and time cost of the cache itself has yet to be quantified.}  In this paper, we present a more comprehensive evaluation \dd{of ER} with several models and tasks\dd{, along with a thorough efficiency analysis}.  We study a simple {\em layer-recycling} ER method that caches the activations from an intermediate layer of a pretrained model, and uses those cached activations as the starting point when the same input sequence is seen again during fine-tuning or inference.  We show that even this simple method yields substantial improvements to throughput at small or no cost to accuracy on average.  Our results provide the strongest evidence to date that ER can be a practically important technique for reducing costs for NLP systems, but as we discuss in~\autoref{sec:discussion}, they also suggest important challenges that must be addressed in future work.

Our contributions are summarized below:

\begin{itemize}
    \item We propose embedding recycling as a method for lowering the computational costs of training and inference for language models, and explore layer recycling with two techniques: standard fine-tuning and parameter-efficient adapters.
    \item Our experiments with eight models across a wide range of tasks show that layer recycling is generally effective.  For the best-performing ER model on our tasks- DeBERTa-XL with adapters, we find that layer recycling nearly matches performance of the original model while providing a 87-91\% speedup at inference time, and greater than 90\% speedup at training time.%, achieving a superior accuracy-efficiency tradeoff to prior QA-specific decomposition techniques \cite{cao2020deformer}.  
    %\mike{speedup based on Luca's BERT results in Table \ref{tab:performance}} \mike{Should we also mention sharing across model architectures here?  Didn't want to get too lengthy but that might be important enough to include}
    \item We explore open challenges for embedding recycling and present questions for future work.
    %\item We show that across a variety of large and compact language models, freezing earlier layers in model architectures can preserve or improve performance on scientific NLP tasks while reducing both training and inference time in half. 
    %\item We demonstrate that by coupling embedding recycling with reduced large language models, we can preserve or improve the performance of the large language models while also gaining new improvements to training and inference times.
    %\item Using embedding recycling, we either match or outperform other competitive models, such as Adapter models and distilled models, but also successfully reuse precomputed embeddings to improve traini   ng and inference times.
    %\item We explore a variety of real-world scenarios for effectively using embedding recycling to avoid recomputations for large document corpora.
\end{itemize}

\section{Related Work}

%Prior studies of transformer fine-tuning have found that shallower layers tend to converge earlier in training than deeper layers \citep{raghu2017svcca,morcos2018insights}, and weights of later layers change more than earlier ones \citet{kovaleva2019revealing}, suggesting that earlier layers tend to extract universal features whereas later layers focus on task-specific modeling.  This indicates that it may not be necessary to fine-tune all the layers of a model to obtain good results.  Indeed,
The embedding recycling technique we investigate is based on findings from prior work suggesting that not all layers of a pretrained transformer are equally important for end-task finetuning.  Shallower layers tend to converge earlier in training than deeper layers \citep{raghu2017svcca,morcos2018insights}, and weights of later layers change more than earlier ones \citep{kovaleva2019revealing}, suggesting that earlier layers tend to extract universal features whereas later layers focus on task-specific modeling.  \citet{Lee2019WhatWE} find that 90\% of fully fine-tuned performance can be reached when fine-tuning only the final quarter of a transformer's layers and leaving the rest frozen.

Several proposed methods vary the number of frozen layers over the course of training, approaching or exceeding the performance of fully fine-tuned models while substantially speeding up the training process \citep{raghu2017svcca, Xiao2019FastDL, Brock2017FreezeOutAT}.  Similar to our approach, some dynamic freezing methods also employed caching mechanisms \citep{Liu2021AutoFreezeAF,he2021pipetransformer}, but the dynamic number of frozen layers means the cache applies only at training time and only for a single task.  In contrast, we cache embeddings from the pretrained model, which can then be reused across multiple downstream tasks and applied at inference time as well.

Other recent studies have sought to improve model inference speed by skipping computations in later layers.  \citet{sajjad2020effect} found that in some cases up to half of the layers can be removed from the model with only a 1-3\% drop in task performance.  %\citet{kumar2019accelerating} considered approximate caching to skip the deeper layers for inputs that produce similar intermediate layer representations.
Early exit strategies have also been proposed, which allow the model to dynamically decide when to skip later layers \citep{cambazoglu2010early,xin2020deebert}.  SkipBERT \citep{wang2022skipbert} combined early exiting with an approach in which cached n-gram embeddings approximate the intermediate activations of new inputs.
\citet{}
%  However, SkipBERT only measures latency (with a batch size of 1), targeting the use case where individual new inputs need to be processed quickly.  In embedding recycling, we focus on the case where an entire cached corpus needs to be processed at once, making throughput more important than small-batch latency.  In addition, SkipBERT's approach is only evaluated on BERT-base and only on sentence-level tasks, whereas we demonstrate the generality of embedding recycling across 8 models and 3 task types.
% - more specific case (stronger assumption)
% - more model sizes
% - measure throughput not latency
% - sequence tagging tasks
\citet{lester2021power} explored prompt-tuning as a parameter-efficient approach for adapting frozen language models without adjusting model weights, conditioning language models with soft prompts to perform downstream tasks.

Precomputing text representations to speed up future processing on the same data is commonly done when creating fixed-size document-level embeddings for use on document-level tasks  \citep{conneau2017supervised,cohan2020specter}; in contrast, we study contextualized {\em token-level} embeddings that can be used for tasks such as named entity recognition (NER) and question answering.  ReadOnce Transformers \citep{lin2021readonce} do consider multi-task variable-length document representations, but do so in a setting where a cached document representation is paired with a query text (such as a question or prompt); the approach is pretrained with QA data and evaluated on QA and summarization, rather than tasks such as text classification or NER where the entire input can be cached.

\citet{du2020general} propose an approach similar to ours that caches general-purpose token-level model representations, trained in a multi-task setting; however, that approach only applies a small MLP to the stored representations and reports a meaningful drop in accuracy (greater than 2\% on average) compared to fully fine-tuned models.  We find that reusing the later layer parameters of a pretrained transformer in addition to the cached activations often enables us to essentially match fully fine-tuned model accuracy while reducing computational cost.%  As noted in our experiments, if we instead use small MLPs from a frozen pretrained encoder, we see large drops in accuracy on our tasks. %\mike{We could mention that we do an analysis of real-world speedup as opposed to just using FLOPS (that turns out to matter since we have some overhead to transfer embeddings), but for now I'm omitting it for brevity}

%\mike{TODO?: mention delta-tuning/parameter-efficient methods, although I'm increasingly unsure if that's actually important to include in this section.}
%Another set of work related to pretrained model reuse is delta tuning \citep{ding2022delta}, a class of methods aiming to train only a small number of parameters to adapt a model for a downstream task.  Delta tuning methods include adapters, bias weights, and soft prompts \citep{houlsby2019parameter, qin2021learning}.

%\mike{TODO?: Maybe mention work on reusing attention, e.g. \citep{xiao2019sharing}}
\citet{Wei2022AFM} combine layer freezing and knowledge distillation to create a multi-task model.  \dd{They do not consider caching activations on persistent storage as we do, but instead re-use activations across tasks at inference time via a branching multi-task model.} They use a two stage process where $12-N$ layers are fine-tuned for each individual task keeping $N$ frozen layers. This is followed by distillation of the $N$ layers for further computational gains. We take advantage of the parameter efficient adapter modules \citep{houlsby2019parameter}, and replace this process with a single step of fine-tuning a frozen base model that has adapters attached only to the deeper layers.

%\arman{added the following paragraph}
%Our work also has connections to prior work on utilizing memory and caching earlier hidden states in a sequence within Transformer models (e.g., \cite{grave2016improving,dai2019transformer,rae-razavi-2020-transformers,wu2022memorizing}). These works generally focus on modeling long range context and caching representations of older history in a sequence. In contrast, our work focuses on caching the representations of the entire sequence and using it for new tasks.  
%\arman{end}

Our work also has connections to work on memory- and retrieval-augmented language modeling.  Prior work on using memory (e.g., \citet{grave2016improving,dai2019transformer,rae-razavi-2020-transformers,wu2022memorizing}) generally focuses on modeling long-range context and caching representations of older history in a sequence, while work on retrieval (e.g., \citet{pmlr-v119-guu20a,karpukhin_etal_2020_dense}) focuses on fetching text from a knowledge base or corpus to serve as additional context.  In both cases, the aim is to use representations of additional text (from earlier in a document or from a knowledge base) to improve modeling of new inputs.  In contrast, our work focuses on caching the representations of an entire sequence to speed up computation for new tasks.

\section{Methods}

\begin{figure}[t]
   \centering
   \includegraphics[width=\linewidth]{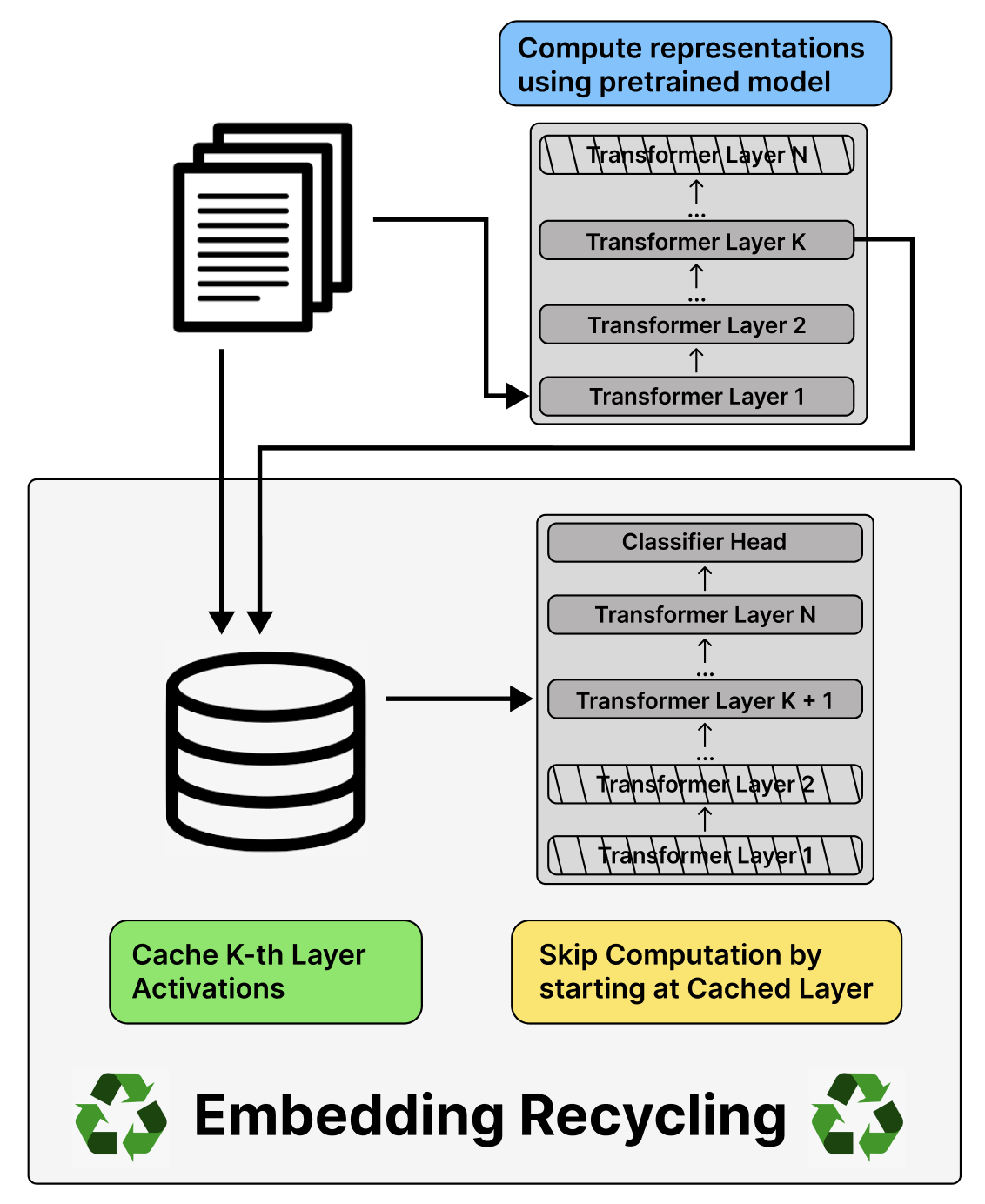}
   \caption{Overview of the embedding recycling approach. In the figure, the K-th layer activations are saved for future fine-tuning on downstream tasks, skipping redundant computations of earlier layers in the transformer model.}
   \label{fig:embedding_recycling}
\end{figure}

In the transformer architecture \citep{vaswani2017attention}, an input sequence $x$ of length $S$ and dimension $d$ is transformed with a function $F: \mathbb{R}^{S\times d} \rightarrow \mathbb{R}^{S\times d}$ defined by the composition of $N$ transformer layers $F^{(1)}, ... , F^{(N)}$ as follows:

\begin{equation} \label{eq:transformer}
    \texttt{F}^{\ell}(x) = \texttt{LN}(\texttt{FF}^{\ell} (\; x') + x') 
\end{equation}

% \begin{equation} \label{eq:self_att}
%     \texttt{SA}^{\ell}(x) = \texttt{FF}^{\ell} (\; x')
% \end{equation}

\begin{equation} \label{eq:self_att}
    \texttt{x'} = \texttt{LN}\Big(\texttt{MH}^{\ell}(x)+x\Big)
\end{equation}

\noindent where $\texttt{LN}$ is a layer normalization \citep{Ba2016LayerN}, $\texttt{FF}$ is a feed forward network, and $\texttt{MH}$ is the self-attention layer that consists of multiple heads and contextualizes the input sequence vector. The output of each layer is used as input to the next layer. 
\begin{equation}
    h^{\ell+1} = F^{\ell}(h^{\ell})
\end{equation}

Our approach is to cache the output representations $h^{k}\in \mathbb{R}^{S\times d}$ at a certain layer $k$ and reuse them for fine-tuning on a new given task. We refer to this process of caching and reusing the output representations of a layer as {\em layer recycling}. This enables us to reduce the size of the transformer model from $N$ layers to $N-k$ layers, reducing the computational cost during fine-tuning and inference. 

Note that the key requirement of layer recycling is that we first need to process the entire data with the transformer model and cache the representations, so that we could later reuse these representations many times during fine-tuning and inference on new tasks. We experiment with two types of layer recycling approaches as explained next.

% Within the transformer architecture, activations are generated at each transformer layer sequentially before being fed into the next layer in the model. In our within-model approach, we use a \textit{source model} to run over a selected corpus before our experiments, caching the activations for each input locally. During the experiments, a \textit{consumer model} then uses the cached activations for fine-tuning and testing. \arman{Cached activations or hidden states?} Both the source model and the consuming model have the same architecture.

We start with a pretrained transformer $F$ (e.g., BERT) consisting of $F^{(1)}, ..., F^{(k)}, ... , F^{(N)}$ layers. During the first epoch of fine-tuning for any given task, we run the transformer over a corpus $\mathcal{C}$ and cache the output representations of layer $k$ for each instance $c$ in $\mathcal{C}$, i.e., $h^{k}_{c\in\mathcal{C}}$. 
However, for every subsequent epoch of fine-tuning using the same transformer model, 
%the same transformer model is then used for fine-tuning on new tasks, however, instead of fine-tuning all the layers, 
we only run and fine-tune the latter $N-k$ layers $F^{(k+1)}, ..., F^{(N)}$.  We can either train all of the weights in the layers (which we refer to as {\em reduced models}), or only train adapter modules added on the layers (discussed below). In either case, for the instance $c$ in the dataset $\mathcal{C}$ we simply retrieve and use the previously cached representation $h^{k}_{c\in\mathcal{C}}$ as input to layer $F^{(k+1)}$. This avoids the extra computation through layers $F^{(1)}, ..., F^{(k)}$ but adds a small cost for retrieving the representation from storage (see \autoref{sec:performance_analysis} for efficiency analysis).  

\subsection{Adapters}

%The layer recycling methods in our experiments use the standard fine-tuning approach for the remaining layers, creating a new reduced model for each new task.  As model size and the number of tasks increases, it could become expensive to store and communicate the layers for each task.  As a proof of concept experiment we evaluate whether recent parameter-efficient techniques can address this issue.  \citet{houlsby2019parameter} introduce adapter modules as a parameter efficient fine-tuning approach which adds a few extra parameters to the transformer model that are trained, leaving the rest of the architecture frozen. \jon{Add motivation for why parameter-efficient storage isn't the focus (since the storage costs of adapters are much smaller than the cost of storing embeddings) and instead the motivation is improved performance}

We evaluate whether combining recycling with Adapter modules~\citep{houlsby2019parameter} can improve performance over fully fine-tuned models. Adapters are typically used to improve the parameter efficiency of fine-tuning and mitigate the storage costs of large language models. They also enable more sample-efficient fine-tuning and can result in improved fine-tuning performance \citep{karimi-mahabadi-etal-2022-prompt}. %Therefore, we investigate whether Adapters can also yield accuracy improvements when used in combination with layer recycling.

Adapter modules contain a down-projection, an up-projection, and a residual connection module: 
$h\leftarrow h+ (f(h\mathbf{W}_{down})\mathbf{W}_{up})$. The adapters are separately inserted after the \texttt{MH} and the \texttt{FF} layers in the transformer architecture (\autoref{eq:self_att}). Further, \citet{Rckl2021AdapterDropOT} experiment with dropping adapters from the lower transformer layers to provide inference time speedup. 
In our experiments, adapters are added to the latter half of transformer layers in the reduced transformer models. 
As in standard layer recycling, the pretrained original transformer $F$ first caches the intermediate activations $h^{k}_{c\in\mathcal{C}}$ for each input in a selected corpus at layer $k$. 
Then the first $k$ layers are removed from the transformer. During fine-tuning, the cached representations are fed as input to the later $N-k$ layers of the transformer, which consist of the frozen transformer layers plus trainable adapter parameters. Thus, we fine-tune only the additional 6-8\% parameters introduced by the adapters.
% since only the adapter parameters need to be stored, which can be cached and retrieved alongside a base model at runtime. 
We refer to learning adapters on all layers as the {\em full adapter} setting and the layer recycling version as the {\em reduced adapter} setting.

\section{Experimental Setup}

We now present our experiments evaluating whether recycled embeddings can be paired with reduced large language models to maintain accuracy while improving training and inference speed.  We explore the effectiveness of embedding recycling across a variety of different tasks, datasets, and transformer models.

\subsection{Models} 
% Based on the multi-layer bidirectional Transformer \citep{vaswani2017attention},
Our full-size models include the encoder transformers  BERT, SciBERT \citep{beltagy2019scibert}, RoBERTa \citep{liu2019roberta}, and DeBERTa \citep{he2020deberta}. We also experiment with the encoder-decoder T5 model \citep{raffel2019exploring}.
%\chrisc{it might be important to specify that T5 is an encoder-decoder model}
We selected these architectures since they are widely-used pretrained transformers across a variety of tasks in different domains. We experiment with multiple sizes of these models, including distilled \citep{sanh2019distilbert,wang2020minilm, wang2021minilmv2}, base, and large variants, to gauge the effectiveness of recycled embeddings with an increase in the network size.

To investigate the effectiveness of layer recycling, we test several reduced models in which we use caching to reduce 50\% of the layers (e.g., caching layer 12 in RoBERTa-large and layer 6 in BERT-base).\footnote{We note that for the encoder-decoder model T5, we consider caching only the middle layer of the {\em encoder}, which means that the speedups for this model will be smaller than (approximately half of) that of the other models we evaluate.  We also consider 25\% and 75\% reduced models in \autoref{sec:experimental_setup}.}  We compare each reduced model to its fully fine-tuned counterpart across the text classification, NER, and QA tasks. The hardware details and hyperparameters for our models are specified in \autoref{sec:experimental_setup}.

% they serve as the current state of the art \arman{These are not state of the art, I'll rephrase}

% Each model has a different number of transformer layers. RoBERTa-Large has 24 transformer layers, BERT-base and SciBERT have 12 layers, and DistilBERT and MiniLM, have 6 transformer layers each. The encoder and decoder side of T5-base (T5-3b) have 12 (24) layers each. T5 models were originally pretrained on a multi-task mixture of unsupervised and supervised tasks, such as question-answering, sentiment analysis, and natural language inference. RoBERTa and BERT, along with distilled versions of the models, were pretrained on the Masked language modeling (MLM) objective using a large corpus of English data. 

\subsection{Datasets}

For our experiments, we focus on three core NLP tasks: text classification, named-entity recognition (NER), and extractive question-answering (QA). Scientific papers, due to their immutable nature, are an especially appropriate target for embedding recycling, so we focus much of our evaluation on the scientific domain.  For text classification, we selected Chemprot \citep{kringelum2016chemprot}, SciCite \citep{cohan-etal-2019-structural}, and SciERC \citep{luan2018multi}. For NER, we used BC5CDR \citep{li2016biocreative}, JNLPBA \citep{collier2004introduction}, and NCBI-Disease \citep{dougan2014ncbi}. For QA, we chose the TriviaQA \citep{joshi2017triviaqa} and SQuAD \citep{rajpurkar2016squad} datasets.

\section{Results}
\label{sec:results}

\subsection{Standard Fine-tuning}

%\mike{Hmm, I have mixed feelings about referring to these models as being "able" or "unable" to do things; seems like a strong claim given that we didn't test configurations exhaustively.  Perhaps "did" or "did not".}

%\mike{Need to define "matches", and may need to adjust claims for "matches or outperforms" depending on how we define it.  See Slack discussion: https://allenai.slack.com/archives/C03D8LCAXMG/p1656716209218279}

\begin{table*}[ht]
\small
\centering
\renewcommand{\arraystretch}{1.1}
\setlength{\tabcolsep}{4.6pt}
\begin{tabular}{@{}lrrrrrrrrrrrrrr@{}}
\toprule
                                                                                            & \multicolumn{2}{c}{\begin{tabular}[c]{@{}c@{}}RoBERTa \\ Large\end{tabular}}                                                                                                            & \multicolumn{2}{c}{(Sci)BERT} & \multicolumn{2}{c}{\begin{tabular}[c]{@{}c@{}}DeBERTa \\ V2 XL \end{tabular}} & \multicolumn{2}{c}{\begin{tabular}[c]{@{}c@{}}T5 \\ Large\end{tabular}}                                                                                                                                                            & \multicolumn{2}{c}{\begin{tabular}[c]{@{}c@{}}MiniLM\\ L6-H768\end{tabular}}                          & \multicolumn{2}{c}{\begin{tabular}[c]{@{}c@{}}MiniLM\\ L6-H384\end{tabular}}                          & \multicolumn{2}{c}{DistilBERT}                    
                                                                                            \\ 
                                                                                            \cmidrule(l{7pt}r{7pt}){2-3} \cmidrule(l{7pt}r{7pt}){4-5} \cmidrule(l{7pt}r{7pt}){6-7} \cmidrule(l{7pt}r{7pt}){8-9} \cmidrule(l{7pt}r{7pt}){10-11} \cmidrule(l{7pt}r{7pt}){12-13}  \cmidrule(l{7pt}r{7pt}){14-15}
Task                                                                               & \multicolumn{1}{c}{\begin{tabular}[c]{@{}c@{}}Rdc\end{tabular}}  & \multicolumn{1}{c}{\begin{tabular}[c]{@{}c@{}}Full\end{tabular}}      & \multicolumn{1}{c}{\begin{tabular}[c]{@{}c@{}}Rdc\end{tabular}}  & \multicolumn{1}{c}{\begin{tabular}[c]{@{}c@{}}Full\end{tabular}}      & \multicolumn{1}{c}{\begin{tabular}[c]{@{}c@{}}Rdc\end{tabular}} & \multicolumn{1}{c}{\begin{tabular}[c]{@{}c@{}}Full\end{tabular}}      & \multicolumn{1}{c}{\begin{tabular}[c]{@{}c@{}}Rdc\end{tabular}} & \multicolumn{1}{c}{\begin{tabular}[c]{@{}c@{}}Full\end{tabular}}  & \multicolumn{1}{c}{\begin{tabular}[c]{@{}c@{}}Rdc\end{tabular}}  & \multicolumn{1}{c}{\begin{tabular}[c]{@{}c@{}}Full\end{tabular}}& \multicolumn{1}{c}{\begin{tabular}[c]{@{}c@{}}Rdc\end{tabular}}  & \multicolumn{1}{c}{\begin{tabular}[c]{@{}c@{}}Full\end{tabular}}& \multicolumn{1}{c}{\begin{tabular}[c]{@{}c@{}}Rdc\end{tabular}}  & \multicolumn{1}{c}{\begin{tabular}[c]{@{}c@{}}Full\end{tabular}} \\ 
\midrule
 
ChemProt                                                                           & {84.3}                                                                                                                                         & \multicolumn{1}{r}{83.9}          & 84.0                                                                                                                                       & \multicolumn{1}{r}{84.0}  & 86.8 & 86.7 & 84.6 & 84.1    & 78.3                                                                      & \multicolumn{1}{r}{{79.3}} & {76.9}                                                             & \multicolumn{1}{r}{74.6}          & {80.3}                                                             & \multicolumn{1}{r}{79.1}                         \\
SciCite                                                                            & 85.0                                                                                                                                                  & \multicolumn{1}{r}{{85.5}} & {86.6}                                                                                                                                         & \multicolumn{1}{r}{86.0} & 85.2 & 84.4 & 86.3 & 84.9   & 84.5                                                                      & \multicolumn{1}{r}{{84.6}} & {83.7}                                                             & \multicolumn{1}{r}{82.8}          & {84.1}                                                             & \multicolumn{1}{r}{84.0}                         \\
SciERC-Rel                                                                        & 80.2                                                                                                                                         & \multicolumn{1}{r}{80.4}          & 76.7                                                                                                                                         & \multicolumn{1}{r}{79.8} & 79.9 & 80.2 & 77.4 & 80.2   & 74.8                                                                      & \multicolumn{1}{r}{{78.2}} & {72.1}                                                             & \multicolumn{1}{r}{68.9}          & {74.9}                                                             & \multicolumn{1}{r}{72.9}                         \\ \hdashline[1pt/1pt]
Classification Avg. & 83.2                                                                                                                                         & \multicolumn{1}{r}{\textbf{83.3}}          & 82.4                                                                                                                                        & \multicolumn{1}{r}{\textbf{83.3}} & \textbf{84.0} & 83.8 & 82.8 & \textbf{83.1}  & 79.2                                                                      & \multicolumn{1}{r}{\textbf{80.7}} & \textbf{77.6}                                                             & \multicolumn{1}{r}{75.4}          & \textbf{79.8}                                                             & \multicolumn{1}{r}{78.7}                         \\ \midrule
bc5cdr                                                                           & 90.0                                                                                                                                                  & \multicolumn{1}{r}{{90.4}} & 90.7                                                                                                                                                  & \multicolumn{1}{r}{{91.3}} & 91.3 & 91.8 & 90.7 & 89.9 & {87.8}                                                             & \multicolumn{1}{r}{87.5}          & 85.9                                                                      & \multicolumn{1}{r}{{88.3}} & 88.3                                                                      & \multicolumn{1}{r}{88.7}                \\
JNLPBA                                                                            & {79.4}                                                                                                                                         & \multicolumn{1}{r}{78.7}          & 78.8                                                                                                                                         & \multicolumn{1}{r}{79.0}  & 78.5 & 78.2 & 79.6 & 80.0 & {77.3}                                                             & \multicolumn{1}{r}{76.9}          & 74.0                                                                      & \multicolumn{1}{r}{{77.2}} & {78.6}                                                             & \multicolumn{1}{r}{78.5}                         \\
NCBI-disease                                                                       & 93.0                                                                                                                                                 & \multicolumn{1}{r}{{93.2}} & {93.4}                                                                                                                                        & \multicolumn{1}{r}{92.9}   & 93.3 & 93.4 & 92.8 & 93.5 & 91.1                                                                      & \multicolumn{1}{r}{{92.1}} & 89.9                                                                      & \multicolumn{1}{r}{{91.7}} & 90.5                                                                      & \multicolumn{1}{r}{91.3}                \\ \hdashline[1pt/1pt]
NER Avg.                & \textbf{87.5}                                                                                                                                         & \multicolumn{1}{r}{87.4}          & \textbf{87.7}                                                                                                                                         & \multicolumn{1}{r}{\textbf{87.7}} & 87.7 & \textbf{87.8} & 87.7 & \textbf{87.8}  & 85.4                                                                      & \multicolumn{1}{r}{\textbf{85.5}} & 83.3                                                                      & \multicolumn{1}{r}{\textbf{85.7}} & 85.8                                                                      & \multicolumn{1}{r}{\textbf{86.2}}     \\ \midrule
TriviaQA                                                          & 78.2                                                                                                                                                  & \multicolumn{1}{r}{{79.8}}  & 67.4                                                                                                                                                  & \multicolumn{1}{r}{{69.1}} & 80.6 & 81.8 & 77.4 & 78.2 & 72.2                                                                      & \multicolumn{1}{r}{{73.8}} & 69.2                                                                      & \multicolumn{1}{r}{{71.0}} & 64.7                                                                      & \multicolumn{1}{r}{66.8}                \\
SQuAD                                                             & 91.8                                                                                                                                                  & \multicolumn{1}{r}{{93.6}} & 87.5                                                                                                                                                 & \multicolumn{1}{r}{{88.5}} & 94.5 & 94.6 & 93.7 & 93.9 & 85.0                                                                      & \multicolumn{1}{r}{{87.0}} & 89.0                                                                      & \multicolumn{1}{r}{{89.6}} & 84.8                                                                      & \multicolumn{1}{r}{85.4}                \\ \hdashline[1pt/1pt]
\begin{tabular}[c]{@{}l@{}}QA Avg.\end{tabular} & 85.0                                                                                                                                                    & \multicolumn{1}{r}{\textbf{86.7}}  & 77.5                                                                                                                                                  & \multicolumn{1}{r}{\textbf{78.8}}  & 87.5 & \textbf{88.2} & 85.5 & \textbf{85.9} & 78.6                                                                      & \multicolumn{1}{r}{\textbf{80.4}} & 79.1                                                                      & \multicolumn{1}{r}{\textbf{80.3}} & 74.8                                                                      & \multicolumn{1}{r}{\textbf{76.1}} \\ \bottomrule          
\end{tabular}
\caption{Test scores of reduced (Rdc) models on the text classification, NER, and QA tasks. \textbf{Bold} indicates the best average F-1 score between the reduced and fully fine-tuned (Full) versions of each model over 10 runs. 
%For each model, the score represents the average macro F-1 score of 10 runs. 
%The F-1 score averages for DeBERTa and T5 were gathered from 5 runs.
%The standard errors for each score are shown in corresponding tables in \autoref{sec:experimental_setup}.  %in \autoref{tab:roberta}, \autoref{tab:bert}, \autoref{tab:distilbert}, \autoref{tab:minilm_768}, and \autoref{tab:minilm_384}.
For the ChemProt dataset, we report the micro F-1 scores instead, following past work \citep{beltagy2019scibert}. 
The reduced BERT-sized models generally offer similar performance to their full counterparts (scoring within 0.2\% when averaged across RoBERTa and SciBERT for the six tasks), and substantially outperform the distilled models. 
%For the QA datasets, we use BERT while we use SciBERT for the text classification and NER datasets since it outperforms BERT on these datasets 
%\citep{beltagy2019scibert}. The reduced models yield a small accuracy drop for QA tasks.
%For the reduced configurations, the half of the transformer layers closest to the input were frozen from fine-tuning. For adapters configurations (Frz+Adpt), we froze all the transformer layers from fine-tuning but added trainable adapters \cite{houlsby2019parameter} to the latter half of the transformer layers furthest from the input.
}
\label{tab:metatable}
\end{table*}

The results for standard fine-tuning of either full or reduced models are shown in \autoref{tab:metatable}.  For the text classification and NER tasks, the reduced BERT-sized and larger models perform similarly to their fine-tuned counterparts on average, and substantially outperform the distilled models.  The reduced distilled models also perform well on those tasks compared to the distilled originals, on average, although there is more variance across models and tasks compared to BERT-sized models. We validate our fully fine-tuned baselines by comparing our results with prior work \citep{beltagy2019scibert}, finding that our scores land within 1.33\% on average and typically score above the previous baselines.
%across the tasks and models For text classification, even with 50\% of the layers frozen, RoBERTa-Large, MiniLM L6-H384, and DistilBERT were able to match or outperform their respective fully fine-tuned configurations. %However, freezing MiniLM L6-H768 layers did not seem to be at par with full fine-tuning.

%For NER, reduced RoBERTa-Large, SciBERT, MiniLM L6-H768, and DistilBERT were generally on par with their respective fully fine-tuned configurations. However, reduced MiniLM L6-H384 did not seem to match the full fine-tuning.

For QA tasks, we found that fully fine-tuning works somewhat better than reduced configurations across all the explored models (\autoref{tab:metatable}). Generally, reduced configurations typically lag by 1 to 2 points in F-1 score.  One possible hypothesis is that the QA datasets are generally much larger than the datasets we used for other tasks (100k-150k examples vs 4k-20k examples for text classification and NER); however, in additional experiments we found that subsampling the QA training sets to 5\% of their original size only increased the gap, suggesting that dataset size does not explain the failure of reduced models on this task.  %As shown in \autoref{sec:experimental_setup}, reducing the model by only 25\% is substantially more effective for QA (average F1 drops by 0.2-0.5 points), which suggests that training layers some of the earlier layers is important for the model to achieve its best performance on the QA datasets, and that future work may consider task-specific choices for which layer to recycle.
We also validate our fully fine-tuned baselines for QA tasks by comparing our results with~\citet{yasunaga2022linkbert}, finding that our scores differ by less than half a point on average. %land within 0.42\% on average.
%Despite the accuracy drop, our QA results compare favorably with previous work.  The QA-specific decomposition techniques in \citep{cao2020deformer} 

Finally, we explored using lightweight multi-layer perceptrons (MLPs) as classifier heads, given their success in prior work. While \cite{du2020general} paired multi-task encoders with 2-layer MLPs, we paired frozen pretrained transformer models with 2-layer MLPs and found that they underperformed trainable layers dramatically, by 26\% on average across the classification and NER tasks. %Furthermore, we found that using a 2-layer MLP instead of a standard linear classifier head in our embedding recycling approach did not meaningfully improve F1-score across text classification, NER, and QA tasks. 
%\jon{Double check phrasing of this paragraph} 
%\dd{I might cut the furthermore sentence here, since it's a somewhat different point (about a design decision within our approach, rather than a comparison vs. Du et al's approach).  You could mention this choice of linear vs. MLP in Sec 4.1 or maybe appendix.}

\subsection{Adapters}

\begin{table*}[ht]
\small
\centering
\renewcommand{\arraystretch}{1.1}
\setlength{\tabcolsep}{4.6pt}
\begin{tabular}{lrrrrrrrrrrrr}
\toprule
\textbf{}           & \multicolumn{3}{c}{\begin{tabular}[c]{@{}c@{}}RoBERTa \\ Large\end{tabular}}                                                                                                                                                            & \multicolumn{3}{c}{(Sci)BERT}                                                                                                                                                               & \multicolumn{3}{c}{\begin{tabular}[c]{@{}c@{}}DeBERTa \\ V2 XL \end{tabular}}                                                                                                                   & \multicolumn{3}{c}{\begin{tabular}[c]{@{}c@{}}T5\\ Large\end{tabular}}                                                                                                                       \\ \cmidrule(l{7pt}r{7pt}){2-4}
\cmidrule(l{7pt}r{7pt}){5-7}
\cmidrule(l{7pt}r{7pt}){8-10} 
\cmidrule(l{7pt}r{7pt}){11-13}
Task                & \multicolumn{1}{c}{\begin{tabular}[c]{@{}c@{}}Rdc +\\ Half\\ Adpt\end{tabular}} & \multicolumn{1}{c}{\begin{tabular}[c]{@{}c@{}}Full\\ Adpt\end{tabular}} & \multicolumn{1}{c}{Full} & \multicolumn{1}{c}{\begin{tabular}[c]{@{}c@{}}Rdc +\\ Half\\ Adpt\end{tabular}} & \multicolumn{1}{c}{\begin{tabular}[c]{@{}c@{}}Full\\ Adpt\end{tabular}} & \multicolumn{1}{c}{Full} & \multicolumn{1}{c}{\begin{tabular}[c]{@{}c@{}}Rdc +\\ Half\\ Adpt\end{tabular}} & \multicolumn{1}{c}{\begin{tabular}[c]{@{}c@{}}Full\\ Adpt\end{tabular}} & \multicolumn{1}{c}{Full} & \multicolumn{1}{c}{\begin{tabular}[c]{@{}c@{}}Rdc +\\ Half\\ Adpt\end{tabular}} & \multicolumn{1}{c}{\begin{tabular}[c]{@{}c@{}}Full\\ Adpt\end{tabular}} & \multicolumn{1}{c}{Full} \\ \midrule
ChemProt            & 84.1                                                                            & 85.2                                                                            & 83.9                     & 84.2                                                                            & 84.9                                                                            & 84.0                     & 87.2                                                                            & 86.5                                                                            & 86.7                     & 84.3                                                                            & 84.9                                                                            & 84.1                     \\
SciCite             & 82.4                                                                            & 82.9                                                                            & 85.5                     & 85.5                                                                            & 84.6                                                                            & 86.0                     & 84.6                                                                            & 85.0                                                                            & 84.4                     & 85.3                                                                            & 84.5                                                                            & 84.9                     \\
SciERC-Rel          & 85.7                                                                            & 85.9                                                                            & 80.4                     & 86.0                                                                            & 85.5                                                                            & 79.8                     & 82.9                                                                            & 82.1                                                                            & 80.2                     & 76.2                                                                            & 75.6                                                                            & 80.2                     \\ \hdashline[1pt/1pt]
Classification Avg. & 84.1                                                                            & \textbf{84.7}                                                                   & 83.3                     & \textbf{85.2}                                                                   & 85.0                                                                              & 83.3                     & \textbf{84.9}                                                                   & 84.6                                                                            & 83.8                     & 81.9                                                                            & 81.7                                                                            & \textbf{83.1}            \\ \midrule
bc5cdr              & 90.0                                                                            & 90.6                                                                            & 90.4                     & 90.0                                                                            & 90.9                                                                            & 91.3                     & 90.7                                                                            & 91.1                                                                            & 91.8                     & 79.9                                                                            & 85.7                                                                            & 89.9                     \\
JNLPBA              & 79.1                                                                            & 79.2                                                                            & 78.7                     & 79.8                                                                            & 78.3                                                                            & 79.0                     & 79.3                                                                            & 79.0                                                                            & 78.2                     & 78.8                                                                            & 79.5                                                                            & 80.0                     \\
NCBI-disease        & 92.8                                                                            & 93.1                                                                            & 93.2                     & 93.1                                                                            & 93.0                                                                            & 92.9                     & 93.3                                                                            & 93.5                                                                            & 93.4                     & 92.1                                                                            & 92.5                                                                            & 93.5                     \\ \hdashline[1pt/1pt]
NER Avg.            & 87.3                                                                            & \textbf{87.6}                                                                   & 87.4                     & 87.6                                                                            & 87.4                                                                            & \textbf{87.7}            & 87.8                                                                            & \textbf{87.9}                                                                   & 87.8                     & 83.6                                                                            & 85.9                                                                            & \textbf{87.8}            \\ \midrule
TriviaQA            & 78.5                                                                            & 79.8                                                                            & 79.8                     & 67.4                                                                            & 68.9                                                                            & 69.1                     & 81.6                                                                            & 82.3                                                                            & 81.8                     & 77.0                                                                            & 77.5                                                                            & 78.2                     \\
SQuAD               & 93.5                                                                            & 93.4                                                                            & 93.6                     & 87.9                                                                            & 87.9                                                                            & 88.5                     & 94.7                                                                            & 93.9                                                                            & 94.6                     & 90.6                                                                            & 91.0                                                                            & 93.9                     \\ \hdashline[1pt/1pt]
QA Avg.             & 86.0                                                                            & 86.6                                                                            & \textbf{86.7}            & 77.6                                                                            & 78.4                                                                            & \textbf{78.8}            & 88.1                                                                            & 88.1                                                                            & \textbf{88.2}            & 83.8                                                                            & {84.3}                                                                   & \textbf{85.9} \\ \bottomrule                   
\end{tabular}
\caption{Test scores of reduced adapter (Rdc + Half Adpt) models on the text classification, NER, and QA tasks. \textbf{Bold} indicates the best average F-1 score between the reduced adapter, full adapter (Full Adpt), and fully fine-tuned (Full) versions of each model over 10 runs. 
%The reduced Adapter-based models are fully frozen reduced model with trainable Adapters added, where the earlier half of the transformer layers were removed. 
%Full Adpt indicates adapters on all transformer layers of a fully frozen model. 
%Score represents the average macro F-1 score of 10 runs. 
%Each score represents the average macro F-1 score of 10 runs for RoBERTa, BERT, and the distilled models. 
%The F-1 score averages for DeBERTa and T5 were gathered from 5 runs.
%The standard errors for each score are shown in corresponding tables in \autoref{sec:experimental_setup}. 
For the ChemProt dataset, we report the micro F-1 scores instead, following past work \citep{beltagy2019scibert}. %For the QA datasets, we use BERT while we use SciBERT for the text classification and NER datasets since it outperforms BERT on these datasets \citep{beltagy2019scibert}.
The reduced, adapter-based transformer models offer similar performance to their full counterparts (scoring within 0.4\% when averaged across RoBERTa, SciBERT, and DeBERTa for the eight tasks), and substantially outperform the distilled models.
%For the reduced configurations, the half of the transformer layers closest to the input were frozen from fine-tuning. For adapters configurations (Frz+Adpt), we froze all the transformer layers from fine-tuning but added trainable adapters \cite{houlsby2019parameter} to the latter half of the transformer layers furthest from the input.
}
\label{tab:adapters}
\end{table*}

Our results for reduced adapter models are shown in \autoref{tab:adapters}.  We see that in general, for all the models except for T5-Large, the adapter-based approaches are superior to standard fine-tuning on our tasks.  Further, layer recycling remains effective with adapters.  Compared to the full adapter baseline, the reduced adapters for RoBERTa-Large, BERT, SciBERT, and DeBERTa models only show a 0.19\% reduction in accuracy. Additionally, compared to the fully fine-tuned baseline, these reduced adapters models have a 0.19-0.23\% reduction in accuracy. 
%\jon{Should we compare to fully finetuned like the abstract or full adapter baselines, as it is now?} \dd{Good question, I'd suggest adding an additional sentence comparing against full fine-tuned; fortunately it makes almost no difference since the numbers are very similar in the two cases?} 
Likewise, in contrast to the full fine-tuning results above, QA accuracy for the top-performing DeBERTa adapter model remains unchanged on average after layer recycling, with the reduced adapter model performing better on one QA task and worse on the other.\footnote{We omit experiments with distilled models, as we found adapters to be ineffective on those models even without embedding recycling, scoring 19.4\% worse on average than full fine-tuning for text classification and NER.} 
%\arman{Consider moving the last sentence to footnote.}

\subsection{GLUE Results}

\begin{table}[ht]
\centering
\small
\begin{tabular}{@{}lcccc@{}}
\toprule
\multirow{2}{*}{\vspace{-1.5em}\textbf{GLUE task}}& \multicolumn{4}{c}{\textbf{DeBERTa V2 XL}\vspace{.2em}} \\
& \multicolumn{1}{l}{\begin{tabular}[c]{@{}c@{}}Rdc +\\ Half Adpt\end{tabular}} & \multicolumn{1}{l}{\begin{tabular}[c]{@{}c@{}}Full\\ Adpt\end{tabular}} & \multicolumn{1}{c}{Rdc} & \multicolumn{1}{c}{Full} \\ \midrule
CoLA & 70.9 & 71.3 & 70.8 & 71.2 \\
SST-2 & 96.9 & 97.1 & 97.1 & 97.4 \\ \hdashline[1pt/1pt]
\begin{tabular}[c]{@{}l@{}}Single\\ Sentence Avg.\end{tabular} & 83.9 & \textbf{84.2} & 84.0 & \textbf{84.3} \\ \midrule
MRPC & 93.9 & 94.0 & 93.4 & 93.9 \\
STS-B & 92.4 & 92.7 & 92.5 & 92.8 \\ \hdashline[1pt/1pt]
\begin{tabular}[c]{@{}l@{}}Similarity and\\ Paraphrase Avg.\end{tabular} & 93.2 & \textbf{93.4} & 93.0 & \textbf{93.4} \\ \midrule
MNLI-m & 91.7 & 92.0 & 91.0 & 91.4 \\
QNLI & 95.0 & 95.1 & 94.1 & 94.8 \\ \hdashline[1pt/1pt]
NLI Avg. & 93.3 & \textbf{93.6} & 92.6 & \textbf{93.1} \\ \midrule 
\end{tabular}
\caption{Test scores of reduced (Rdc) and reduced adapter (Rdc + Half Adpt) models on GLUE for DeBERTa V2 XL. \textbf{Bold} indicates the best average score between the reduced and fully fine-tuned (Full) versions for the standard and adapter-based configurations. Each score is averaged over 5 runs.
%Our results for MNLI correspond to MNLI-matched. 
We report the scores using the standard GLUE metric for each corresponding task.}
\label{tab:glue}
\end{table}

For our best-performing model DeBERTa v2 XL, we also provide further experiments on datasets from the GLUE benchmark \citep{wang2018glue}, to allow easier comparison against speedup techniques from previous work. We present results on the CoLA, SST-2, MRPC, STS-B, MNLI, and QNLI tasks from GLUE. For our experiments, we tried both our standard reduced models and our reduced adapter models. We found that embedding recycling was successful across the GLUE tasks, with an average accuracy drop of 0.3 points in return for a significant increase in both training and inference time as outlined in \autoref{tab:performance_training} and \autoref{tab:performance}.  We note that due to the high computational cost of these experiments, we take existing hyperparameter settings from previous work that worked well for the full models, and also use these for reduced models. Further hyperparameter optimization of the reduced models might improve performance.

\subsection{Efficiency Analysis}
\label{sec:performance_analysis}

\begin{table}[t]
\footnotesize
\centering
\small
\renewcommand{\arraystretch}{1.1}
\setlength{\tabcolsep}{3.6pt}
\begin{tabular}{ccccc}
\toprule
& \multicolumn{3}{c}{
    \renewcommand{\arraystretch}{1}
    \begin{tabular}[c]{@{}c@{}}
        \textbf{Inference Time}\\
        {\footnotesize\color{gray}(Speedup over Baseline)}\vspace{.1em}
    \end{tabular}
} & \\
\multicolumn{1}{c}{\multirow{3}{*}{\textbf{Model}}} &
    \multicolumn{1}{c}{\multirow{3}{*}{\textbf{Baseline}}} &
    \multicolumn{2}{c}{\textbf{Recycling}\vspace{.1em}} &
    \multicolumn{1}{c}{
        \multirow{3}{*}{
        \renewcommand{\arraystretch}{1}
        \begin{tabular}[c]{@{}c@{}}
            \textbf{Avg. F1} \\
            \textbf{diff} when \\
            \textbf{recycling} \vspace{.1em}
        \end{tabular}
    }} \\
\multicolumn{1}{c}{} &
    \multicolumn{1}{c}{} &
    \multicolumn{1}{c}{
        \renewcommand{\arraystretch}{.9}
        \begin{tabular}[c]{@{}c@{}}
            FP32 \\ cache
        \end{tabular}
    } &
    \multicolumn{1}{c}{
        \renewcommand{\arraystretch}{.9}
        \begin{tabular}[c]{@{}c@{}}
            FP16 \\ cache
        \end{tabular}
    } \\
    % \multicolumn{1}{c}{NR vs FP32} &
    % \multicolumn{1}{c}{NR vs FP16} \\
\midrule
\multicolumn{5}{c}{\textit{NVIDIA A10G}\vspace{.1em}} \\
\multicolumn{1}{c}{
    \renewcommand{\arraystretch}{.9}
    \begin{tabular}[c]{@{}c@{}}
        MiniLM\\{\scriptsize \textsc{L6-H384}}\vspace{.1em}
    \end{tabular}
} & 183 \textit{ms} &
    \multicolumn{1}{c}{
        \renewcommand{\arraystretch}{1}
        \begin{tabular}[c]{@{}c@{}}
            154 \textit{ms} \\ {\footnotesize\color{gray} (+21\%)}\vspace{.1em}
        \end{tabular}
    } &
    \multicolumn{1}{c}{
        \renewcommand{\arraystretch}{1}
        \begin{tabular}[c]{@{}c@{}}
            123 \textit{ms} \\ {\footnotesize\color{gray} (+67\%)}\vspace{.1em}
        \end{tabular}
    } & $-0.2$ \\  % & +21\% & +67\% & -0.2 \\
\multicolumn{1}{c}{
    \renewcommand{\arraystretch}{.9}
    \begin{tabular}[c]{@{}c@{}}
        MiniLM\\{\scriptsize \textsc{L6-H768}}\vspace{.1em}
    \end{tabular}
} & 325 \textit{ms} &
    \multicolumn{1}{c}{
        \renewcommand{\arraystretch}{1}
        \begin{tabular}[c]{@{}c@{}}
            201 \textit{ms} \\ {\footnotesize\color{gray} (+56\%)}\vspace{.1em}
        \end{tabular}
    } &
    \multicolumn{1}{c}{
        \renewcommand{\arraystretch}{1}
        \begin{tabular}[c]{@{}c@{}}
            195 \textit{ms} \\ {\footnotesize\color{gray} (+66\%)}\vspace{.1em}
        \end{tabular}
    } & $-0.4$ \\  % & +56\% & +66\% & -0.4 \\
\multicolumn{1}{c}{
    \renewcommand{\arraystretch}{.9}
    \begin{tabular}[c]{@{}c@{}}
        BERT\\ {\scriptsize\textsc{Base}}\vspace{.1em}
    \end{tabular}
} & 647 \textit{ms} &
    \multicolumn{1}{c}{
        \renewcommand{\arraystretch}{1}
        \begin{tabular}[c]{@{}c@{}}
            351 \textit{ms} \\ {\footnotesize\color{gray} (+84\%)}\vspace{.1em}
        \end{tabular}
    } &
    \multicolumn{1}{c}{
        \renewcommand{\arraystretch}{1}
        \begin{tabular}[c]{@{}c@{}}
            343 \textit{ms} \\ {\footnotesize\color{gray} (+88\%)}\vspace{.1em}
        \end{tabular}
    } & $-0.3$ \\  % & +84\% & +88\% & -0.3 \\
\multicolumn{1}{c}{
    \renewcommand{\arraystretch}{.9}
    \begin{tabular}[c]{@{}c@{}}
        BERT\\ {\scriptsize\textsc{Large}}\vspace{.1em}
    \end{tabular}
} & 1943 \textit{ms} &
    \multicolumn{1}{c}{
        \renewcommand{\arraystretch}{1}
        \begin{tabular}[c]{@{}c@{}}
            1066 \textit{ms} \\ {\footnotesize\color{gray} (+86\%)}\vspace{.1em}
        \end{tabular}
    } &
    \multicolumn{1}{c}{
        \renewcommand{\arraystretch}{1}
        \begin{tabular}[c]{@{}c@{}}
            1004 \textit{ms} \\ {\footnotesize\color{gray} (+93\%)}\vspace{.1em}
        \end{tabular}
    } & $-0.2$ \\  % & +86\% & +93\% & -0.2 \\
\multicolumn{1}{c}{
    \renewcommand{\arraystretch}{.9}
    \begin{tabular}[c]{@{}c@{}}
        DeBERTa\\ {\scriptsize\textsc{V2-XLarge}}\vspace{.1em}
    \end{tabular}
} & 1914 \textit{ms} &
    \multicolumn{1}{c}{
        \renewcommand{\arraystretch}{1}
        \begin{tabular}[c]{@{}c@{}}
            1010 \textit{ms} \\ {\footnotesize\color{gray} (+89\%)}\vspace{.1em}
        \end{tabular}
    } &
    \multicolumn{1}{c}{
        \renewcommand{\arraystretch}{1}
        \begin{tabular}[c]{@{}c@{}}
            985 \textit{ms} \\ {\footnotesize\color{gray} (+94\%)}\vspace{.1em}
        \end{tabular}
    } & $-0.1$ \\ % & +89\% & +94\% & -0.1 \\
\midrule
\multicolumn{5}{c}{\textit{NVIDIA A6000}\vspace{.1em}} \\
\multicolumn{1}{c}{
    \renewcommand{\arraystretch}{.9}
    \begin{tabular}[c]{@{}c@{}}
        MiniLM\\{\scriptsize \textsc{L6-H384}}\vspace{.1em}
    \end{tabular}
} & 123 \textit{ms} &
    \multicolumn{1}{c}{
        \renewcommand{\arraystretch}{1}
        \begin{tabular}[c]{@{}c@{}}
            105 \textit{ms} \\ {\footnotesize\color{gray} (+18\%)}\vspace{.1em}
        \end{tabular}
    } &
    \multicolumn{1}{c}{
        \renewcommand{\arraystretch}{1}
        \begin{tabular}[c]{@{}c@{}}
            100 \textit{ms} \\ {\footnotesize\color{gray} (+23\%)}\vspace{.1em}
        \end{tabular}
    } & $-0.2$ \\  % & +18\% & +23\% & -0.2 \\
\multicolumn{1}{c}{
    \renewcommand{\arraystretch}{.9}
    \begin{tabular}[c]{@{}c@{}}
        MiniLM\\{\scriptsize \textsc{L6-H768}}\vspace{.4em}
    \end{tabular}
} & 208 \textit{ms} &
    \multicolumn{1}{c}{
        \renewcommand{\arraystretch}{1}
        \begin{tabular}[c]{@{}c@{}}
            161 \textit{ms} \\ {\footnotesize\color{gray} (+29\%)}\vspace{.1em}
        \end{tabular}
    } &
    \multicolumn{1}{c}{
        \renewcommand{\arraystretch}{1}
        \begin{tabular}[c]{@{}c@{}}
            150 \textit{ms} \\ {\footnotesize\color{gray} (+38\%)}\vspace{.1em}
        \end{tabular}
    } & $-0.4$ \\  % & +29\% & +38\% & -0.4 \\
\multicolumn{1}{c}{
    \renewcommand{\arraystretch}{.9}
    \begin{tabular}[c]{@{}c@{}}
        BERT\\ {\scriptsize\textsc{Base}}\vspace{.4em}
    \end{tabular}
} & 416 \textit{ms} &
    \multicolumn{1}{c}{
        \renewcommand{\arraystretch}{1}
        \begin{tabular}[c]{@{}c@{}}
            269 \textit{ms} \\ {\footnotesize\color{gray} (+55\%)}\vspace{.1em}
        \end{tabular}
    } &
    \multicolumn{1}{c}{
        \renewcommand{\arraystretch}{1}
        \begin{tabular}[c]{@{}c@{}}
            245 \textit{ms} \\ {\footnotesize\color{gray} (+59\%)}\vspace{.1em}
        \end{tabular}
    } & $-0.3$ \\  % & +55\% & +59\% & -0.3 \\
\multicolumn{1}{c}{
    \renewcommand{\arraystretch}{.9}
    \begin{tabular}[c]{@{}c@{}}
        BERT\\ {\scriptsize\textsc{Large}}\vspace{.4em}
    \end{tabular}
} & 1235 \textit{ms} &
    \multicolumn{1}{c}{
        \renewcommand{\arraystretch}{1}
        \begin{tabular}[c]{@{}c@{}}
            662 \textit{ms} \\ {\footnotesize\color{gray} (+86\%)}\vspace{.1em}
        \end{tabular}
    } &
    \multicolumn{1}{c}{
        \renewcommand{\arraystretch}{1}
        \begin{tabular}[c]{@{}c@{}}
            643 \textit{ms} \\ {\footnotesize\color{gray} (+92\%)}\vspace{.1em}
        \end{tabular}
    } & $-0.2$ \\  % & +86\% & +92\% & -0.2 \\
\multicolumn{1}{c}{
    \renewcommand{\arraystretch}{.9}
    \begin{tabular}[c]{@{}c@{}}
        DeBERTa\\ {\scriptsize\textsc{V2-XLarge}}\vspace{.4em}
    \end{tabular}
} & 1430 \textit{ms} &
    \multicolumn{1}{c}{
        \renewcommand{\arraystretch}{1}
        \begin{tabular}[c]{@{}c@{}}
            777 \textit{ms} \\ {\footnotesize\color{gray} (+84\%)}\vspace{.1em}
        \end{tabular}
    } &
    \multicolumn{1}{c}{
        \renewcommand{\arraystretch}{1}
        \begin{tabular}[c]{@{}c@{}}
            758 \textit{ms} \\ {\footnotesize\color{gray} (+89\%)}\vspace{.1em}
        \end{tabular}
    } & $-0.1$ \\ % & +84\% & +89\% & -0.1 \\
\bottomrule
\end{tabular}
\caption{
Average \textbf{inference} runtime comparison (in ms/batch, averaged over 7 runs) between vanilla encoders and models that cache embeddings on disk.
% We assume the cache is precomputed (see \autoref{sec:performance_analysis}).
For all runs, cache the middle layer of the encoder.  We assume the cache is already precomputed when calculating timings; thus, maximum speedup is 100\%.
Overall, the larger the model, the higher the speedup from re-using representations.
Further, accelerators with fewer execution units (A10G) benefit more from recycling embeddings.
Finally, using half precision for embeddings improves speed up across the board, while halving storage size.
% We also found that half-precision has a negligible effect on F1-scores at inference if you originally train the models using full-precision.
% In the rightmost column, we included the average F1 loss from using embedding recycling across our tasks.
}
\label{tab:performance}
\end{table}

To estimate the real-world benefit of recycling embeddings for different tasks, we provide a minimal PyTorch implementation of embedding recycling. 
This implementation and the following results correspond to both the standard layer recycling approach and the adapter-based layer recycling approach since they follow parallel processes for gradient descent during training and computations during inference, despite the additional 6-8\% of parameters added by the trainable adapters. To show that training times do not differ substantially, we also measured the training time the transformer models take to converge to their optimal weights. We found both approaches take approximately the same time to complete training (\autoref{tab:training_times}).

We evaluated the impact of recycling embeddings on four different architectures and two different hardware platforms. 
For models, we considered two efficient transformer models (MiniLMv2 \citep{wang2020minilm,wang2021minilmv2} models with $l=6$ layers and embeddings of size $h=384$ and $h=768$), 
two medium sized models ($\text{BERT}_{\textsc{Base}}$, $l=12$, $h=768$; $\text{BERT}_{\textsc{Large}}$, $l=24$, $h=1024$), 
and a large model ($\text{DeBERTa}_{\textsc{V2-XLarge}}$, $l=24$, $h=1536$).
We evaluated embeddings on a efficiency-oriented AI accelerator (NVIDIA A10G), as well as on a high-performance GPU (NVIDIA A6000).

We controlled for differences among tasks considered in tables~\autoref{tab:metatable},~\ref{tab:adapters},~and~\ref{tab:glue}, such as length of sequences and number of samples, by simulating a sequence classification task on \textsc{Qasper} \citep{dasigi2021qasper}, which includes the full-text of over a thousand academic manuscripts.\footnote{Because the bulk of computation for a transformer model is done in its encoder and not in the task-specific heads, inference time is similar regardless of whether the model is used for sequence classification, tagging, or question answering.}
We run all models with a fixed batch size of 128.
For all models, we reduce exactly half of their layers by recycling, which results in a maximum theoretical speed-up of 100\%.  
A run over the corpus consists of 335 batches, and we average results over seven runs.

~\autoref{tab:performance} shows the results of caching embeddings to recycle on disk.
Overall, we found that all models benefit from embedding recycling, achieving an average speedup ranging from 18 to 86\%. 
Unsurprisingly, larger models benefit more from recycling than smaller ones; this is due to the fact that loading embeddings cached on disk adds a small latency penalty to a model run, which is more noticeable in the case of smaller models. 
For example, we achieve an 84\% speedup when running $\text{BERT}_{\textsc{Base}}$ with embedding recycling on an A10G GPU, which is roughly equivalent to the latency of a $\text{MiniLM}_{\text{L6-H768}}$ model without recycling (351 vs 325 ms per batch on average);
this result would us allow to run more accurate models while maintaining the efficiency of shallower architectures. 

%We note that in our inference-time speedup measurements, we assume that the cache is already pre-computed. 
%This corresponds to our target setting in which new tasks and models are executed over the same text that has been processed previously.  
%Because the cost to write the cache to disk is approximately equal to a single inference pass over the corpus, if we perform $t$ total inference passes for different models/tasks given the same pre-computed cache, the total amortized speedup will be approximately $\frac{t}{t+1}$ of the values we report in \autoref{tab:performance}. 
%Thus, as the number of inference passes to be run increases, the total amoritized speedup including the cost to write the cache will approach the values reported in the table.

\begin{table*}[t]
\footnotesize
\centering
\small
\renewcommand{\arraystretch}{1.1}
\setlength{\tabcolsep}{4.3pt}
\begin{tabular}{cccccccc} 
\toprule
\multicolumn{1}{c}{\multirow{3}{*}{\textbf{Model}}} & \multicolumn{4}{c}{\textbf{Training} (ms/batch, amortized over \textbf{ 6 epochs})} & \multicolumn{3}{c}{\textbf{Speedup}} \\
 & \begin{tabular}[c]{@{}c@{}}\textbf{No}\\\textbf{Recycling} (NR)\end{tabular} & \begin{tabular}[c]{@{}c@{}}\textbf{Model}\\\textbf{Frozen} (F)\end{tabular} & \multicolumn{1}{l}{\begin{tabular}[c]{@{}c@{}}\textbf{Saving} \textplus \\\textbf{Recycling} (SR)\end{tabular}} & \multicolumn{1}{l}{\begin{tabular}[c]{@{}c@{}}\textbf{Only} \\\textbf{Recycling} (R)\end{tabular}} & \begin{tabular}[c]{@{}c@{}}\textbf{NR} vs\\ \textbf{SR}\end{tabular} &  \begin{tabular}[c]{@{}c@{}}\textbf{F} vs\\ \textbf{SR}\end{tabular} &  \begin{tabular}[c]{@{}c@{}}\textbf{NR} vs\\ \textbf{R}\end{tabular} \\
\midrule
\multicolumn{8}{c}{NVIDIA A10G} \\
$\text{MiniLM}_{384}$ & 51 $\pm$ 1 & 30 $\pm$ 1 & 32 $\pm$ 6 & 25 $\pm$ 4 & +59\% & -7\% & +104\% \\
$\text{MiniLM}_{768}$ & 90 $\pm$ 4 & ~56 $\pm$ 1 & 50 $\pm$ 4 & 45 $\pm$ 3 & +80\% & +12\% & +100\% \\
$\text{BERT}_{\textsc{Base}}$ & 173 $\pm$ 2 & 112 $\pm$ 1 & 90 $\pm$ 4 & 87 $\pm$ 3 & +92\% & +24\% & +99\% \\
$\text{BERT}_{\textsc{Large}}$ & 347 $\pm$ 1 & 246 $\pm$ 1 & 181 $\pm$ 2 & 176 $\pm$ 2 & +92\% & +36\% & +97\% \\
$\text{DeBERTa}_{\textsc{XLarge}}$ & 380 $\pm$ 2 & 286 $\pm$ 1 & 199 $\pm$ 1 & 194 $\pm$ 1 & +91\% & +44\% & +96\% \\
\midrule
\multicolumn{8}{c}{NVIDIA A6000} \\
$\text{MiniLM}_{384}$ & 41 $\pm$ 1 & 24 $\pm$ 1 & 26 $\pm$ 5 & 22 $\pm$ 3 & +55\% & -8\% & +81\% \\
$\text{MiniLM}_{768}$ & 61 $\pm$ 1 & 38 $\pm$ 1 & 40 $\pm$ 5 & 34 $\pm$ 3 & +52\% & -5\% & +82\% \\
$\text{BERT}_{\textsc{Base}}$ & 117 $\pm$ 1 & 78 $\pm$ 1 & 60 $\pm$ 3 & 58 $\pm$ 2 & +94\% & +30\% & +102\% \\
$\text{BERT}_{\textsc{Large}}$ & 326 $\pm$ 2 & 212 $\pm$~1 & 167 $\pm$ 2 & 161 $\pm$ 1 & +96\% & +26\% & +103\% \\
$\text{DeBERTa}_{\textsc{XLarge}}$ & 359 $\pm$ 2 & 250 $\pm$~1 & 184 $\pm$ 1 & 178 $\pm$ 1 & +95\% & +35\% & +102\% \\
\bottomrule
\end{tabular}
\caption{
Average \textbf{training} runtime comparison (in ms per batch, $\pm$ stdev over 7 runs) between vanilla encoders and models that cache embeddings on disk. 
For all runs, we cache the middle layer of the encoder; thus, theoretical speedup is 100\%. 
Time per batch is amortized over 6 epochs ($2,000$ steps), the lowest number to convergence over all datasets (c.r.f. Table~\ref{tab:training_times}). 
We present results in four settings: no recycling (NR), freezing \textonehalf~of the layers during training (F), 1 training epoch during which embeddings are saved to disk followed by 5 epochs where recycling is enabled (SR), and 6 epochs where embeddings are already saved (R).
Overall, we found that embedding recycling speeds up training even when embeddings need to be cached to disk during the first pass.
Compared to freezing, saving and recycling improves training time for all but  MiniLM models (F vs SR).
}
\label{tab:performance_training}
\end{table*}

\autoref{tab:performance} also includes results when storing embeddings using half precision (that is, cache embeddings in FP16 rather FP32). 
The smaller embeddings lead to improvements for all models and hardware, ranging from $+8\%$ to $+46\%$. 
Further, it has no impact on performance, as it changes predicted scores by at most $10^{-3}$ across all tasks evaluated in this work.

We also note that less capable hardware benefits more from caching embeddings. 
For example, $\text{BERT}_{\textsc{Base}}$ achieves a speedup of $84\%$ on an A10G GPU, while on A6000, the speedup is a more modest $55\%$. 
This is an expected result: fewer and slower execution cores/accelerator memory impact overall model latency.
Further, we note that, despite the smaller relative gains, the more powerful GPU is always faster in absolute terms compared with the less capable one.

It is important to note that these gaps from  maximum achievable speedup are only observed when performing \textit{inference}; 
for \textit{training}, we observe almost perfect speed-up for all models and hardware configurations except for the smaller MiniLM models. 
For example, $\text{BERT}_{\textsc{Base}}$ requires $17.38 \pm 1.32\text{ ms/batch}$\footnote{When training, we use a batch size of 16} without recycling, compared to $8.67 \pm 2.18\text{ ms/batch}$ when recycling.
Even when considering the additional time to cache embeddings to disk during the first pass, embedding recycling still achieves close to optimum speedup on all models except MiniLMs, where its gains hover between 52\% and 82\% (``NR vs SR'' column in \autoref{tab:performance_training}).
When training for just 6 epochs (or roughly $2,000$ steps), recycling embeddings is faster than simply freezing half of the parameters for all models but MiniLM (``F vs SR'' column in \autoref{tab:performance_training}); this is due to the relatively higher cost of caching layers to disk in case of smaller models.
In these cases, we empirically found that recycling achieves faster training time than freezing after 12 epochs or $4,000$ training steps; 
since smaller models typically require more epochs to converge, we conclude that recycling is generally preferable to partially freezing a model during training.
For $\text{BERT}_{\textsc{Base}}$ and larger models, embedding recycling is also more efficient than layer freezing, providing a $+20\%$ to $+45\%$ speed-up after just 6 training epochs. 
%Full training results are reported in ~\autoref{sec:perf_training} and~\autoref{tab:performance_training} of the appendix.

We also benchmarked the storage requirements of recycling embeddings. 
For a sequence of 512 tokens and a hidden model dimension of 768, caching embeddings requires 1.6 MB with 32-bit precision or 0.8 MB with 16-bit precision. 
This translates to 15.5 MB per paper in \textsc{Qasper} (papers are, on average 4,884 WordPiece tokens long).
Weighing the storage cost and compute savings of ER, we find that it is cost-effective in cloud environments only if the corpus is reprocessed several times per month, but is cost-effective on local hardware even with infrequent (yearly) corpus reprocessing (details in \autoref{appendix:cost-analysis} of the appendix).

\section{Discussion and Future Work}
\label{sec:discussion}

%For text classification tasks, we found that both frozen RoBERTa and SciBERT with adapters on the latter half of their transformer layers outperform their respective fully fine-tuned versions substantially on SciERC-Rel and slightly on ChemProt, but underperform on SciCite.  The differences between the adapter-augmented and fully-finetuned models are smaller for the NER datasets, and have approximately equal performance on average.

% \subsection{Directions for Future Work}

Our experiments raise several questions \dd{and suggest multiple avenues for future work, including:}
% and embedding recycling contains a much larger space of potential techniques than those we investigate here.  Future work could proceed along various lines, including:
\begin{itemize}[leftmargin=*, itemindent=7pt]
    \item Our layer recycling strategy is a straightforward ER approach, but previous work has suggested that weighted pooling across layers can perform better %in probing tasks
    compared to any single layer in many cases \citep{liu-etal-2019-linguistic,du2020general}.  
    Recycling pooled activations may offer improved results.  
    What is the best way to capture and store the syntactic and semantic knowledge encoded in the activations of a model for later recycling?
    \item As noted in the previous section, naive storage methods for ER can be cost-prohibitive in some settings, and finding ways to mitigate this cost (e.g., by compressing the stored activations) will be important for making ER broadly applicable.
    \item Our experiments show that the right recycling approach may be task-specific and model-specific.  For example, with standard fine-tuning as shown in \autoref{tab:roberta}, caching layer 12 in RoBERTa-large is most effective for NER and text classification, whereas it is not effective for QA (but layer 6 performs much better).  Which embeddings to retrieve and recycle for a task, and the right architecture (e.g. number of layers) to use when consuming the recycled embeddings, represents a large decision space.  Methods that can help practitioners automatically choose among public or private shared embedding sets and associated model designs, given their task and objectives for accuracy and computational cost, may be important to make ER an effective practical tool.
    \item We present results with encoder-only and encoder-decoder models, on classification tasks.  Determining whether the approach is effective for generative tasks and autoregressive models is an important question for future work.
    \item
    %Other than recycling, a variety of other inference-speedup techniques exist for large neural models.  
    While we show that ER can be effective when coupled with distillation, whether other techniques like quantization and early exiting remain effective in combination with ER is an open question. %One  straightforward and potentially promising direction would be to combine the "late start" of layer recycling ER with early exit techniques \citep{xin2020deebert}. 
    \item We focus on the setting where the exact same text, at the length of a full document, is being reused for multiple tasks.  
    In practice, we may often perform a task on text that is {\em similar} to but not exactly the same as one for which we have cached embeddings (e.g., a Wikipedia page that has been revised).
    Further, even a completely new document will have similarities and overlapped spans with previously processed ones.
    Studying ER in these settings, e.g. through a combination of layer recycling and the SkipBERT approach which can apply to unseen passages via cached n-grams \citep{wang2022skipbert}, is an area of future work.
    \item Finally, it is possible to explore cross-model embedding recycling. We attempted a straightforward implementation of such approach by using recycling embeddings from a larger model into a smaller consumer model. However, the results did not show improvements (Appx. \ref{sec:appendix_cross_model_er}).
    % However, using rich contextualized embeddings from a large model to help power many smaller downstream task models is an important setting for ER, since it provides a powerful way to amortize the expense of running a large model. 
    Developing and evaluating new approaches for this setting is an important item for future work.
\end{itemize}

\section{Conclusion}

We have presented embedding recycling, a general technique for reusing previous activations of neural language models to improve the efficiency of future training and inference.  We show how a simple technique of caching a layer of activations in a pretrained model is effective.  We validate our approach in experiments across fourteen tasks and eight model architectures.  We find that recycling typically has small or no impacts to accuracy on average, but does yield substantial throughput increases demonstrated through a careful efficiency analysis.  We also discuss several open challenges for future work.

\section{Limitations}

As discussed in detail in our future work section, several advances are important to make embedding recycling a broadly applicable practical technique.  In addition, the techniques we evaluate primarily benefit transformer language models run on GPU-based architectures with rapid storage, components which are not available to all NLP researchers and practitioners.
Our experiments demonstrate positive results with one representative embedding recycling technique, but do not directly evaluate all recycling variants proposed earlier in the literature.  %Furthermore, while large storage is not necessary to reproduce our experimental results, deploying embedding recycling across various models and datasets can require larger storage units and the funds to sustain them.
Finally, the datasets used in our experiments were in English, a high-resource language with robust pretrained models which may benefit embedding recycling.  Future work should expand on the applicability of embedding recycling by using non-English datasets in lower-resource settings to determine the breadth of its applicability.

\section*{Acknowledgments}
\dd{
This work was supported in part by NSF Convergence Accelerator Grant OIA-2033558.  We thank Chris Coleman for helpful discussions.
}

\bibliography{tacl2021}
\bibliographystyle{acl_natbib}

\appendix

%\section{Appendix}
\section{Experimental Setup and Additional Results}
\label{sec:experimental_setup}

\subsection{Fine-tuning Transformer Models}

The candidate transformer models are fine-tuned using configurations suggested by \citet{devlin2018bert}, \citet{ding2022delta} and \citet{houlsby2019parameter}. For text classification, we feed the final hidden state of the \texttt{[CLS]} token into a linear classification layer. For NER and QA, we feed the final hidden states of each token into a linear classification layer with a softmax output.

For all of the models, we apply a dropout of 0.1 to the transformer outputs and optimize for cross entropy loss using Adam \citep{kingma2014adam}. We employ a batch size of 32 across all tasks. We fine-tune using early stopping with a patience of 10, using a validation set for calculating loss for each epoch. We use a linear warmup followed by linear decay for training \citep{howard2018universal}, testing the following learning rate options: 1e-3, 2e-3, 1e-4, 2e-4, 1e-5, 2e-5, 5e-5, and 5e-6. For the text classification and NER datasets, we select the best performing learning rate for each transformer model on the development set and report the corresponding test results. For the QA datasets, we select the best performing learning rate for each transformer model on the training set and report the corresponding results on the validation set. Additionally, for the adapter modules used in certain model configurations, we test bottleneck dimensions as part of our hyperparameter search: 24, 64, and 256. 

\subsection{Adapter-based Models}
Here, we used frozen RoBERTa-Large \citep{liu2019roberta}, SciBERT \citep{beltagy2019scibert}, and BERT models but added adapter modules \citep{houlsby2019parameter} only on the latter half of the transformer layers. Only the adapters and the linear classifier attached to the model output were fine-tuned for the text classification, NER, and QA tasks. 

We found that the best hyperparameter configuration was generally a bottleneck dimension of 256 and a learning rate of either 1e-4 or 2e-4.

\subsection{Cross-model Embedding Reuse}
\label{sec:appendix_cross_model_er}

\begin{table*}[t]
\centering
\small
\begin{tabular}{llrrrr}
\toprule
                                                         & \textbf{}                                     & \multicolumn{1}{l}{\textbf{\begin{tabular}[c]{@{}l@{}}RoBERTa-Large\\ + MiniLM L6-H768\end{tabular}}} & \multicolumn{1}{l}{\textbf{MiniLM L6-H768}} & \multicolumn{1}{l}{\textbf{\begin{tabular}[c]{@{}l@{}}BERT + \\ DistilBERT\end{tabular}}} & \multicolumn{1}{l}{\textbf{DistilBERT}} \\ \midrule
\textbf{Chemprot}                                        & Micro F-1                                     & 78.9 (0.3)                                                                                            & \multicolumn{1}{r}{{79.3 (0.3)}}    & 77.8 (0.4)                                                                                & {79.1 (0.5)}                     \\
\textbf{}                                                & Macro F-1                                     & 52.2 (0.2)                                                                                            & \multicolumn{1}{r}{{52.6 (0.4)}}    & 51.2 (0.5)                                                                                & {52.6 (0.3)}                     \\
\textbf{SciCite}                                         & Micro F-1                                     & 85.2 (0.3)                                                                                            & \multicolumn{1}{r}{{86.0 (0.2)}}    & {85.7 (0.1)}                                                                       & 85.5 (0.1)                              \\
\textbf{}                                                & Macro F-1                                     & 83.8 (0.3)                                                                                            & \multicolumn{1}{r}{{84.6 (0.2)}}    & {84.2 (0.1)}                                                                       & 84.0 (0.1)                              \\
\textbf{SciERC-Rel}                                      & Micro F-1                                     & 85.1 (0.4)                                                                                            & \multicolumn{1}{r}{{86.3 (0.2)}}    & {83.8 (0.2)}                                                                       & 83.5 (0.4)                              \\
                                                         & Macro F-1                                     & 76.2 (0.8)                                                                                            & \multicolumn{1}{r}{78.2 (0.6)}    & {73.6 (0.6)}                                                                       & 72.9 (0.7)                              \\ \midrule
\multicolumn{2}{l}{\textbf{\begin{tabular}[c]{@{}l@{}}Text Classification\\ Average Score\end{tabular}}} & 76.9                                                                                                  & \multicolumn{1}{r}{\textbf{77.8}}          & 76.0                                                                                      & \textbf{76.3}  \\ \bottomrule                        
\end{tabular}
\caption{Cross-Model Recycling Results for RoBERTa+MiniLM-L6H768 and BERT+DistilBERT configurations. \textbf{Bold} indicates the best average score between the cross-model recycling and fully finetuned versions of each model. Each score represents the average score of 10 runs, with the standard errors for each score in parentheses.}
\label{tab:crossmodel}
\end{table*}

An alternative to re-using cached activation from a pre-trained model (\autoref{sec:results}), is to cache activations from a more expensive, larger model and re-using them in downstream cheaper models. The goal here is to improve accuracy by using more powerful contextual embeddings. Overall, a straightforward implementation of this strategy did not offer improvements, as described below.

We experiment with reusing precomputed embeddings from one source model $F$ in a consumer model $F'$ that has a different size but the same tokenization vocabulary. 
The activations of the \emph{final} transformer layer $h^{N}_{c\in\mathcal{C}}$ are stored for each input $c$ from corpus $C$. During the fine-tuning of the consumer model $F'$, these stored activations are transformed through a learned 2-layer MLP with ReLU activation\footnote{We found that MLP achieved better performance compared with a single linear layer on dev set.} and added to the input embeddings of $F'$.
We tried two frameworks for pairing large language model embeddings with compact models: $F$=Roberta-large $\rightarrow$ $F'$=MiniLM-6L-H768 and $F$=BERT-base $\rightarrow$ $F'$=DistilBERT.

Overall, as shown in \autoref{tab:crossmodel} the larger model's contextual representations do not improve the smaller model's accuracy; in fact adding them decreases the average F1 score by 0.3-0.9 points.  %In future work, we hope to explore different approaches for cross-model recycling.

\subsection{Efficiency of Embedding Recycling when Training}
\label{sec:perf_training}

For training, we observe almost perfect speed-up for all models and hardware configuration, barring MiniLM models on the machine equipped with a A6000 GPU (``NR vs R'' column in \autoref{tab:performance_training}).  
For example, $\text{BERT}_{\textsc{Base}}$ requires $17.38 \pm 1.32\text{ ms/batch}$\footnote{When training, we use a batch size of 16} without recycling, compared to $8.67 \pm 2.18\text{ ms/batch}$ when recycling.
Even when considering the additional time to cache embeddings to disk during the first pass, embedding recycling still achieves close to optimum speedup on all models except MiniLMs, where its gains hover between 52\% and 82\% (``NR vs SR'' column in \autoref{tab:performance_training}).
When training for just 6 epochs (or roughly $2,000$ steps), recycling embeddings is faster than simply freezing half of the parameters for all models but MiniLM (``F vs SR'' column in \autoref{tab:performance_training}); this is due to the relatively higher cost of caching layers to disk in case of smaller models.
In these cases, we empirically found that recycling achieves faster training time than freezing after 12 epochs or $4,000$ training steps; 
since smaller models typically require more epochs to converge, we conclude that recycling is generally preferable to partially freezing a model during training.

\subsection{Embedding Pre-fetching while Recycling}
Storing embeddings on NVMe drives, while fast, introduce additional latency compared to RAM. 
For example, $\text{BERT}_{\textsc{Base}}$ achieves an average latency of $351 \pm 1 \text{ ms/batch}$ when caching on disk ($84\%$ speedup), compared to just $334 \pm 1 \text{ ms/batch}$ when using memory ($94\%$ speedup). 
This is due to the fact that, while embeddings are being loaded from disk, the hardware accelerator responsible for executing the rest of the model sits idle. 
To reduce the impact of this latency penalty, our implementation supports \textit{pre-fetching} of future embeddings: when processing a sequence of inputs, such as sentences in a manuscript, it loads embeddings for tokens ahead of the sequence inference is currently being run on.
This optimization reduces the time accelerators wait for data to be available for inference; 
for example, in the case of $\text{BERT}_{\textsc{Base}}$ on A10G, disabling pre-fetching raised inference inference time to $374 \pm 1 \text{ ms/batch}$ (vs $351 \pm 1 \text{ ms/batch}$ with pre-fetching). 
Therefore in this section, all results are reported with prefetching enabled.

\subsection{Software and Hardware}

For implementation, we use the v4.19 version of the Transformers library \citep{wolf2019huggingface}, the v0.4 version of the OpenDelta library \citep{ding2022delta}, and the v1.11 version of the Pytorch library \citep{paszke2019pytorch}. We conduct our experiments using NVIDIA RTX A6000 GPUs and NVIDIA A10G GPUs with CUDA v11.5.

\subsection{Considerations in Selecting Hardware for Proof-of-Concept Recycling Experiments}

We ran our proof-of-concept implementation on an AWS Cloud instance\footnote{ \texttt{g5.2xlarge} instance with 8 cores and 32 GB of RAM.} equipped with an NVIDIA A10G accelerator, and on a NVIDIA A6000 within an on-premise server\footnote{Intel-based system with 128 cores and 512 GB of RAM.}. 
The former contains fewer execution units (72 vs 84), fewer tensor cores (288 vs 336), slower memory (600 vs 768 GB/s), and slower boost clock (1800 MHz vs 1695 MHz). However, it is much more efficient, being rated at 150W (compare with A6000's 300W power target). 
Therefore, the NVIDIA A10G accelerator presents a more realistic platform for embedding recycling, since it is more suitable for cost-efficient large-scale model deployments.
Both machines are equipped with PCIe NVMe drives, which we use to cache embeddings to recycle. 

%\vspace{18cm}

%\section{Additional results}
%\label{sec:appendix_results}

\subsection{Cost-effectiveness of Embedding Recycling}
\label{appendix:cost-analysis}
In this section we attempt to estimate how cost-effective embedding recycling is for inference in a real-world setting.  While this depends heavily on use-case-specific assumptions, we consider two typical settings as proofs-of-concept, one using cloud computing and one using local hardware.

There are four main factors that affect the cost-benefit ratio of embedding recycling: (1) compute cost, (2) storage cost, (3) model architecture, and (4) frequency of corpus reprocessing (i.e., how often the cached embeddings will be used).  Compute costs are challenging to estimate for a locally-owned hardware setting due to many hidden cost factors beyond the GPUs (cooling, electrical costs, server to house the GPUs, etc) and so we use AWS EC2 cloud GPU prices as a cost estimate for both cloud and local hardware.  In particular, we consider a \texttt{g5.12xlarge} instance with 4 $\times$ A10G GPUs at 5.67 \$/hr.

Storage costs are easier to estimate for local hardware than compute costs, and local storage can be significantly cheaper because embedding recycling does not require the availability and durability guarantees provided by cloud solutions (the cache is accessed infrequently and can always be recomputed if it is lost).  Therefore, we consider both a cloud storage solution (AWS S3 one-zone infrequent access, at 0.01 \$/GB/month) and a local storage solution.  For local storage, we consider current consumer-grade hard drive prices at approximately 16.9 \$/TB based on data from Amazon and Newegg, and assume a lifespan of 6 years based on data from Backblaze.\footnote{https://www.backblaze.com/blog/how-long-do-disk-drives-last/}  This results in an average cost of 0.23 \$/TB/month over the life of the drive.  Finally, we note that AWS does not charge for data transfer between S3 and EC2 within a region, so we can ignore data transfer costs in this calculation.

The frequency of corpus reprocessing is highly variable, so we report results in terms of the minimum reprocessing frequency that would be necessary for embedding recycling to be cost-effective.  For all models we assume each input is 512 tokens and the cache is stored with FP16 precision.

\begin{table}[t]
    \centering
    \begin{tabular}{lcc}
        \toprule
        {\bf Model} & {\bf Cloud} & {\bf Local} \\
        \midrule
        $\text{MiniLM}_{384}$              & 0.05 & 2.2 \\
        $\text{MiniLM}_{768}$              & 0.05 & 2.4 \\
        $\text{BERT}_{\textsc{Base}}$      & 0.13 & 5.6 \\
        $\text{BERT}_{\textsc{Large}}$     & 0.30 & 12.9 \\
        $\text{DeBERTa}_{\textsc{XLarge}}$ & 0.20 & 8.5 \\
        \bottomrule
    \end{tabular}
    \caption{Minimum reprocessing frequency (in months) needed in order for embedding recycling to be cost-effective in various model and hardware configurations.}
    \label{tab:er-cost-benefit}
\end{table}

%\mike{Hmm, maybe these numbers should go in a table}
%Under our assumptions, we find that embedding recycling is cost-effective in a cloud setting if the corpus is reprocessed very frequently.  Specifically, the corpus would need to be reprocessed 19.8 times per month with $\text{MiniLM}_{384}$, 18.3 times per month with $\text{MiniLM}_{768}$, 7.8 times per month with $\text{BERT}_{\textsc{Base}}$, 3.4 times per month with ($\text{BERT}_{\textsc{Large}}$), or 5.1 times per month with $\text{DeBERTa}_{\textsc{XLarge}}$ to justify the storage costs.  However, with local hardware the calculation is more favorable; embedding recycling with $\text{BERT}_{\textsc{Large}}$ would be worthwhile even if the corpus were only reprocessed once per year (12.9 months).  The benefit varies across models, with a minimum reprocessing frequency of 2.2 months for $\text{MiniLM}_{384}$, 2.4 months for $\text{MiniLM}_{768}$, 5.6 months for $\text{BERT}_{\textsc{Base}}$, and 8.5 months for $\text{DeBERTa}_{\textsc{XLarge}}$.
\autoref{tab:er-cost-benefit} shows the minimum reprocessing frequency needed for embedding recycling to be cost effective for our models on cloud and local hardware.  Under our assumptions, we find that embedding recycling is cost-effective in a cloud setting only if the corpus is reprocessed very frequently (several times per month).  This may be realistic in some use cases, such as when a large team is working with the same corpus and developing many new models, or if new training data arrives frequently and the model developer wants to continually update and re-deploy it.

With local hardware the calculation is much more favorable; embedding recycling with $\text{BERT}_{\textsc{Large}}$ would be worthwhile even if the corpus were only reprocessed once per year.

We note that embedding recycling could become substantially more cost effective with further development.  In this work we did not explore ways to reduce storage costs, such as quantization or compression.  In addition, while our experiments only considered sequence lengths of 512 tokens, for many full-text document corpora it is desirable to use a much longer sequence length to fit the whole document into a model at once.  Because the computational cost of transformers generally scales superlinearly with input length (but storage cost scales only linearly), embedding recycling will be more effective as the sequence length grows.

\begin{table*}[t]
\centering
\small
\setlength{\tabcolsep}{2.3pt}
\begin{tabular}{llrrrrrr}
\toprule
                                                          &                                              & \multicolumn{1}{l}{}                                                                      & \multicolumn{1}{l}{}                                                                      & \multicolumn{4}{c}{\textbf{RoBERTa-Large}}                                                                                                                                                                                                                                                                                                                                                   \\ \midrule
                                                          & \textbf{}                                    & \multicolumn{1}{l}{\textbf{\begin{tabular}[c]{@{}l@{}}Reduced +\\ Half Adpt\end{tabular}}} & \multicolumn{1}{l}{\textbf{\begin{tabular}[c]{@{}l@{}}Full \\ Adapters\end{tabular}}} & \multicolumn{1}{l}{\textbf{\begin{tabular}[c]{@{}l@{}}6 Layers \\ Reduced \end{tabular}}} & \multicolumn{1}{l}{\textbf{\begin{tabular}[c]{@{}l@{}}12 Layers \\ Reduced \end{tabular}}} & \multicolumn{1}{l}{\textbf{\begin{tabular}[c]{@{}l@{}}18 Layers \\ Reduced \end{tabular}}} & \multicolumn{1}{l}{\textbf{\begin{tabular}[c]{@{}l@{}}Fully \\ Finetuned\end{tabular}}} \\ \midrule
\textbf{ChemProt}                                         & Micro F-1                                    & 84.1 (0.4)                                                                                & \textbf{85.2 (0.3)}                                                                       & 84.2 (0.3)                                                                                     & 84.3 (0.2)                                                                                      & 82.0 (0.2)                                                                                      & 83.9 (0.3)                                                                              \\
\textbf{}                                                 & Macro F-1                                    & \textbf{60.8 (0.7)}                                                                       & 57.5 (0.7)                                                                                & 56.4 (0.4)                                                                                     & 56.5 (0.3)                                                                                      & 54.5 (0.5)                                                                                      & 56.5 (0.4)                                                                              \\
\textbf{SciCite}                                          & Micro F-1                                    & 85.2 (0.3)                                                                                & 85.6 (0.5)                                                                                & 86.2 (0.2)                                                                                     & 86.2 (0.2)                                                                                      & 86.2 (0.2)                                                                                      & \textbf{86.8 (0.2)}                                                                     \\
\textbf{}                                                 & Macro F-1                                    & 82.4 (0.4)                                                                                & 82.9 (0.6)                                                                                & 84.9 (0.2)                                                                                     & 85.0 (0.2)                                                                                      & 85.0 (0.2)                                                                                      & \textbf{85.5 (0.2)}                                                                     \\
\textbf{SciERC-Rel}                                       & Micro F-1                                    & 89.0 (0.5)                                                                                & \textbf{89.3 (0.6)}                                                                       & 87.1 (0.4)                                                                                     & 86.8 (0.4)                                                                                      & 86.1 (0.2)                                                                                      & 87.3 (0.4)                                                                              \\
\textbf{}                                                 & Macro F-1                                    & 85.7 (0.7)                                                                                & \textbf{85.9 (0.9)}                                                                       & 79.4 (0.7)                                                                                     & 80.2 (0.8)                                                                                      & 76.2 (0.4)                                                                                      & 80.4 (0.6)                                                                              \\ \midrule
\multicolumn{2}{l}{\textbf{\begin{tabular}[c]{@{}l@{}}Text Classification\\ Average Score\end{tabular}}} & \textbf{81.2}                                                                             & 81.1                                                                                      & 79.7                                                                                           & 79.8                                                                                            & 78.3                                                                                            & 80.1                                                                                    \\ \midrule
\textbf{bc5cdr}                                           & Micro F-1                                    & 97.4 (0.0)                                                                                & \textbf{97.6 (0.0)}                                                                       & 97.2 (0.3)                                                                                     & 97.4 (0.0)                                                                                      & 97.3 (0.0)                                                                                      & 97.5 (0.0)                                                                              \\
\textbf{}                                                 & Macro F-1                                    & 90.0 (0.0)                                                                                & \textbf{90.6 (0.0)}                                                                       & 89.0 (1.2)                                                                                     & 90.0 (0.0)                                                                                      & 89.5 (0.1)                                                                                      & 90.4 (0.1)                                                                              \\
\textbf{JNLPBA}                                           & Micro F-1                                    & 93.8 (0.0)                                                                                & 93.8 (0.0)                                                                                & 93.8 (0.0)                                                                                     & \textbf{93.9 (0.0)}                                                                             & 93.7 (0.0)                                                                                      & 93.7 (0.1)                                                                              \\
\textbf{}                                                 & Macro F-1                                    & 79.1 (0.1)                                                                                & 79.2 (0.2)                                                                                & 79.3 (0.1)                                                                                     & \textbf{79.4 (0.1)}                                                                             & 79.0 (0.1)                                                                                      & 78.7 (0.3)                                                                              \\
\textbf{NCBI-disease}                                     & Micro F-1                                    & 98.5 (0.0)                                                                                & \textbf{98.6 (0.0)}                                                                       & 98.5 (0.0)                                                                                     & 98.5 (0.0)                                                                                      & 98.4 (0.0)                                                                                      & \textbf{98.6 (0.0)}                                                                     \\
\textbf{}                                                 & Macro F-1                                    & 92.8 (0.1)                                                                                & 93.1 (0.1)                                                                                & 93.0 (0.1)                                                                                     & 93.0 (0.1)                                                                                      & 92.4 (0.1)                                                                                      & \textbf{93.2 (0.1)}                                                                     \\ \midrule
\multicolumn{2}{l}{\textbf{\begin{tabular}[c]{@{}l@{}}NER Average \\ Score\end{tabular}}}                & 91.9                                                                                      & \textbf{92.1}                                                                             & 91.8                                                                                           & 92.0                                                                                            & 91.7                                                                                            & 92.0                                                                                    \\ \midrule
\textbf{TriviaQA}                                         & Micro F-1                                    & 75.3 (0.1)                                                                                & \textbf{76.8 (0.2)}                                                                       & 76.6 (0.2)                                                                                     & 75.1 (0.1)                                                                                      & 70.8 (0.1)                                                                                      & 76.7 (0.1)                                                                              \\
\textbf{}                                                 & Macro F-1                                    & 78.5 (0.1)                                                                                & \textbf{79.8 (0.1)}                                                                       & 79.7 (0.2)                                                                                     & 78.2 (0.1)                                                                                      & 73.8 (0.1)                                                                                      & \textbf{79.8 (0.1)}                                                                     \\
\textbf{SQuAD}                                            & Micro F-1                                    & 87.0 (0.1)                                                                                & 86.7 (0.0)                                                                                & 86.2 (0.0)                                                                                     & 84.7 (0.0)                                                                                      & 79.3 (0.0)                                                                                      & \textbf{87.4 (0.0)}                                                                     \\
                                                          & Macro F-1                                    & 93.5 (0.1)                                                                                & 93.4 (0.0)                                                                                & 92.8 (0.0)                                                                                     & 91.8 (0.0)                                                                                      & 87.8 (0.0)                                                                                      & \textbf{93.6 (0.0)}                                                                     \\ \midrule
\multicolumn{2}{l}{\textbf{\begin{tabular}[c]{@{}l@{}}QA Average\\ Score\end{tabular}}}                  & 83.6                                                                                      & 84.1                                                                                      & 83.8                                                                                           & 82.4                                                                                            & 77.9                                                                                            & \textbf{84.3} \\ \bottomrule                                                                         
\end{tabular}
\caption{RoBERTa Results for Reduced Models. \textbf{Bold} indicates the best average score between the standard reduced, adapter-based reduced, and fully fine-tuned versions of each model. \textbf{Reduced + Half Adpt} indicates adapters on the transformer layers of a fully frozen reduced model, where the earlier half of transformer layers were removed and their activations cached. \textbf{Full Adapters} indicates adapters on all transformer layers of a fully frozen model. Each score represents the average score of 10 runs, with the standard errors for each score in parentheses.}
\label{tab:roberta}
\end{table*}

\begin{table*}[t]
\centering
\small
\setlength{\tabcolsep}{2.3pt}
\begin{tabular}{llrrrrrr}
\toprule
                                                       &                                                       & \multicolumn{1}{l}{}                                                                      & \multicolumn{1}{l}{}                                                                      & \multicolumn{4}{c}{\textbf{SciBERT}}                                                                                                                                                                                                                                                                                                                                                       \\ \midrule
                                                       &                                 & \multicolumn{1}{l}{\textbf{\begin{tabular}[c]{@{}l@{}}Reduced +\\ Half Adpt\end{tabular}}} & \multicolumn{1}{l}{\textbf{\begin{tabular}[c]{@{}l@{}}Full \\ Adapters\end{tabular}}} & \multicolumn{1}{l}{\textbf{\begin{tabular}[c]{@{}l@{}}3 Layers \\ Reduced \end{tabular}}} & \multicolumn{1}{l}{\textbf{\begin{tabular}[c]{@{}l@{}}6 Layers \\ Reduced \end{tabular}}} & \multicolumn{1}{l}{\textbf{\begin{tabular}[c]{@{}l@{}}9 Layers \\ Reduced \end{tabular}}} & \multicolumn{1}{l}{\textbf{\begin{tabular}[c]{@{}l@{}}Fully \\ Finetuned\end{tabular}}} \\ \midrule
\textbf{ChemProt}                                      & Micro F-1                                             & 84.2 (0.3)                                                                                & \textbf{84.9 (0.4)}                                                                       & 83.8 (0.4)                                                                                     & 84.0 (0.2)                                                                                     & 81.9 (0.2)                                                                                     & 84.0 (0.3)                                                                              \\
\textbf{}                                              & Macro F-1                                             & 56.9 (0.8)                                                                                & 54.8 (0.4)                                                                                & 56.5 (0.5)                                                                                     & \textbf{57.0 (0.3)}                                                                            & 54.3 (0.3)                                                                                     & 56.3 (0.4)                                                                              \\
\textbf{SciCite}                                       & Micro F-1                                             & 86.6 (0.2)                                                                                & 85.8 (0.1)                                                                                & 87.1 (0.1)                                                                                     & \textbf{87.6 (0.1)}                                                                            & 87.4 (0.1)                                                                                     & 87.1 (0.2)                                                                              \\
\textbf{}                                              & Macro F-1                                             & 85.5 (0.3)                                                                                & 84.6 (0.1)                                                                                & 86.1 (0.1)                                                                                     & \textbf{86.6 (0.1)}                                                                            & 86.2 (0.1)                                                                                     & 86.0 (0.2)                                                                              \\
\textbf{SciERC-Rel}                                    & Micro F-1                                             & \textbf{89.4 (0.4)}                                                                       & 88.5 (0.6)                                                                                & 86.6 (0.3)                                                                                     & 86.1 (0.2)                                                                                     & 85.4 (0.2)                                                                                     & 86.3 (0.2)                                                                              \\
\textbf{}                                              & Macro F-1                                             & \textbf{86.0 (0.7)}                                                                       & 85.5 (0.6)                                                                                & 77.6 (0.5)                                                                                     & 76.7 (0.3)                                                                                     & 76.2 (0.4)                                                                                     & 79.8 (0.5)                                                                              \\ \midrule
\multicolumn{2}{l}{\textbf{\begin{tabular}[c]{@{}l@{}}Text Classification\\ Average Performance\end{tabular}}} & \textbf{81.4}                                                                             & 80.7                                                                                      & 79.6                                                                                           & 79.7                                                                                           & 78.6                                                                                           & 79.9                                                                                    \\ \midrule
\textbf{bc5cdr}                                        & Micro F-1                                             & 97.5 (0.0)                                                                                & \textbf{97.7 (0.1)}                                                                       & 97.7 (0.0)                                                                                     & 97.6 (0.0)                                                                                     & 97.5 (0.0)                                                                                     & \textbf{97.7 (0.0)}                                                                     \\
\textbf{}                                              & Macro F-1                                             & 90.0 (0.0)                                                                                & 90.9 (0.1)                                                                                & 91.0 (0.1)                                                                                     & 90.7 (0.0)                                                                                     & 90.2 (0.1)                                                                                     & \textbf{91.3 (0.0)}                                                                     \\
\textbf{JNLPBA}                                        & Micro F-1                                             & \textbf{94.0 (0.0)}                                                                       & 93.5 (0.0)                                                                                & 93.6 (0.1)                                                                                     & 93.7 (0.1)                                                                                     & 93.8 (0.0)                                                                                     & 93.6 (0.1)                                                                              \\
\textbf{}                                              & Macro F-1                                             & \textbf{79.8 (0.0)}                                                                       & 78.3 (0.2)                                                                                & 78.6 (0.4)                                                                                     & 78.8 (0.2)                                                                                     & 79.0 (0.1)                                                                                     & 79.0 (0.2)                                                                              \\
\textbf{NCBI-disease}                                  & Micro F-1                                             & \textbf{98.6 (0.0)}                                                                       & 98.5 (0.0)                                                                                & 98.5 (0.0)                                                                                     & \textbf{98.6 (0.0)}                                                                            & 98.5 (0.0)                                                                                     & 98.5 (0.0)                                                                              \\
\textbf{}                                              & Macro F-1                                             & 93.1 (0.1)                                                                                & 93.0 (0.1)                                                                                & 92.9 (0.1)                                                                                     & \textbf{93.4 (0.1)}                                                                            & 93.1 (0.1)                                                                                     & 92.9 (0.1)                                                                              \\ \midrule
\multicolumn{2}{l}{\textbf{\begin{tabular}[c]{@{}l@{}}NER Average\\ Perforamcne\end{tabular}}}                 & \textbf{92.2}                                                                             & 92.0                                                                                      & 92                                                                                             & 92.1                                                                                           & 92                                                                                             & \textbf{92.2}  \\ \bottomrule                                                                        
\end{tabular}
\caption{SciBERT text classification and NER results for Reduced Models. \textbf{Bold} indicates the best average score between the standard reduced, adapter-based reduced, and fully fine-tuned versions of each model. \textbf{Reduced + Half Adpt} indicates adapters on the transformer layers of a fully frozen reduced model, where the earlier half of transformer layers were removed and their activations cached. \textbf{Full Adapters} indicates adapters on all transformer layers of a fully frozen model. Each score represents the average score of 10 runs, with the standard errors for each score in parentheses. QA tasks are not included since SciBERT was pretrained for scientific datasets.}
\label{tab:scibert}
\end{table*}

\begin{table*}[htp]
\centering
\small
\setlength{\tabcolsep}{2.3pt}
\begin{tabular}{llrrrrrr}
\toprule
                                                &                                        & \multicolumn{6}{c}{\textbf{BERT}}                                                                                                                                                                                                                                                                                                                                                                                                                                                                                                                                                  \\ \midrule
                                                &                                        & \multicolumn{1}{l}{\textbf{\begin{tabular}[c]{@{}l@{}}Reduced +\\ Half Adpt\end{tabular}}} & \multicolumn{1}{l}{\textbf{\begin{tabular}[c]{@{}l@{}}Full \\ Adapters\end{tabular}}} & \multicolumn{1}{l}{\textbf{\begin{tabular}[c]{@{}l@{}}3 Layers \\ Reduced \end{tabular}}} & \multicolumn{1}{l}{\textbf{\begin{tabular}[c]{@{}l@{}}6 Layers \\ Reduced \end{tabular}}} & \multicolumn{1}{l}{\textbf{\begin{tabular}[c]{@{}l@{}}9 Layers \\ Reduced \end{tabular}}} & \multicolumn{1}{l}{\textbf{\begin{tabular}[c]{@{}l@{}}Fully \\ Finetuned\end{tabular}}} \\ \midrule
\textbf{TriviaQA}                               & Micro F-1                              & 63.9 (0.5)                                                                                & 65.5 (0.1)                                                                                & 65.7 (0.1)                                                                                     & 64.1 (0.2)                                                                                     & 61.4 (0.1)                                                                                     & \textbf{66.0 (0.1)}                                                                     \\
\textbf{}                                       & Macro F-1                              & 67.4 (0.5)                                                                                & 68.9 (0.1)                                                                                & 68.9 (0.1)                                                                                     & 67.4 (0.1)                                                                                     & 64.8 (0.1)                                                                                     & \textbf{69.1 (0.1)}                                                                     \\
\textbf{SQuAD}                                  & Micro F-1                              & 80.2 (0.1)                                                                                & 80.2 (0.0)                                                                                & 80.8 (0.1)                                                                                     & 79.5 (0.1)                                                                                     & 75.4 (0.1)                                                                                     & \textbf{81.1 (0.1)}                                                                     \\
\textbf{}                                       & Macro F-1                              & 87.9 (0.1)                                                                                & 87.9 (0.0)                                                                                & 88.4 (0.1)                                                                                     & 87.5 (0.1)                                                                                     & 84.8 (0.1)                                                                                     & \textbf{88.5 (0.0)}                                                                     \\ \midrule
\multicolumn{2}{l}{\textbf{\begin{tabular}[c]{@{}l@{}}QA Average\\ Scores\end{tabular}}} & 74.9                                                                                      & 75.6                                                                                      & 76.0                                                                                           & 74.6                                                                                           & 71.6                                                                                           & \textbf{76.2}        \\ \midrule                                                                  
\end{tabular}
\caption{BERT QA Results for Reduced Models. \textbf{Bold} indicates the best average score between the standard reduced, adapter-based reduced, and fully fine-tuned versions of each model. \textbf{Reduced + Half Adpt} indicates adapters on the transformer layers of a fully frozen reduced model, where the earlier half of transformer layers were removed and their activations cached. \textbf{Full Adapters} indicates adapters on all transformer layers of a fully frozen model. Each score represents the average score of 10 runs, with the standard errors for each score in parentheses.}
\label{tab:bert}
\end{table*}

\begin{table*}[htp]
\centering
\small
\setlength{\tabcolsep}{2.3pt}
\begin{tabular}{llrrrrrr}
\toprule
                                                          &                                              & \multicolumn{1}{l}{}                                                                          & \multicolumn{1}{l}{}                                                                           & \multicolumn{4}{c}{\textbf{DeBERTaV2 XL}}                                                                                                                                                                                                                                                                                                                                                    \\ \midrule
                                                          & \textbf{}                                    & \multicolumn{1}{l}{\textbf{\begin{tabular}[c]{@{}l@{}}Reduced +\\ Half Adpt\end{tabular}}} & \multicolumn{1}{l}{\textbf{\begin{tabular}[c]{@{}l@{}}Full \\ Adapters\end{tabular}}} & \multicolumn{1}{l}{\textbf{\begin{tabular}[c]{@{}l@{}}6 Layers \\ Reduced \end{tabular}}} & \multicolumn{1}{l}{\textbf{\begin{tabular}[c]{@{}l@{}}12 Layers \\ Reduced \end{tabular}}} & \multicolumn{1}{l}{\textbf{\begin{tabular}[c]{@{}l@{}}18 Layers \\ Reduced \end{tabular}}} & \multicolumn{1}{l}{\textbf{\begin{tabular}[c]{@{}l@{}}Fully \\ Finetuned\end{tabular}}} \\ \midrule
\textbf{ChemProt}                                         & Micro F-1                                    & \textbf{87.2 (0.1)}                                                                           & 86.5 (0.2)                                                                                     & \textbf{87.2 (0.2)}                                                                            & 86.8 (0.4)                                                                                      & 86.4 (0.2)                                                                                      & 86.7 (0.9)                                                                              \\
\textbf{}                                                 & Macro F-1                                    & 56.7 (0.5)                                                                                    & 55.6 (0.6)                                                                                     & \textbf{59.6 (0.2)}                                                                            & 59.5 (0.5)                                                                                      & 59.2 (0.3)                                                                                      & 59.0 (1.1)                                                                              \\
\textbf{SciCite}                                          & Micro F-1                                    & 85.8 (0.4)                                                                                    & \textbf{86.4 (0.4)}                                                                            & 86.0 (0.1)                                                                                     & 86.3 (0.2)                                                                                      & 86.2 (0.3)                                                                                      & 85.9 (0.2)                                                                              \\
\textbf{}                                                 & Macro F-1                                    & 84.6 (0.4)                                                                                    & 85.0 (0.5)                                                                                     & 84.6 (0.1)                                                                                     & \textbf{85.2 (0.1)}                                                                             & 85.0 (0.3)                                                                                      & 84.4 (0.2)                                                                              \\
\textbf{SciERC-Rel}                                       & Micro F-1                                    & \textbf{88.6 (0.5)}                                                                           & 88.0 (0.4)                                                                                     & 88.3 (0.2)                                                                                     & 87.5 (0.1)                                                                                      & 86.6 (0.3)                                                                                      & 88.0 (0.4)                                                                              \\
\textbf{}                                                 & Macro F-1                                    & \textbf{82.9 (0.8)}                                                                           & 82.1 (0.8)                                                                                     & 80.5 (0.5)                                                                                     & 79.9 (0.3)                                                                                      & 78.0 (0.4)                                                                                      & 80.2 (0.5)                                                                              \\ \midrule
\multicolumn{2}{l}{\textbf{\begin{tabular}[c]{@{}l@{}}Text Classification\\ Average Score\end{tabular}}} & \textbf{81.0}                                                                                & 80.6                                                                                         & \textbf{81.0}                                                                                 & 80.9                                                                                          & 80.2                                                                                           & 80.7                                                                                   \\ \midrule
\textbf{bc5cdr}                                           & Micro F-1                                    & 97.6 (0.0)                                                                                    & 97.7 (0.0)                                                                                     & 97.4 (0.3)                                                                                     & 97.7 (0.0)                                                                                      & 97.6 (0.0)                                                                                      & \textbf{97.9 (0.0)}                                                                     \\
\textbf{}                                                 & Macro F-1                                    & 90.7 (0.1)                                                                                    & 91.1 (0.1)                                                                                     & 89.5 (1.4)                                                                                     & 91.3 (0.0)                                                                                      & 90.9 (0.0)                                                                                      & \textbf{91.8 (0.1)}                                                                     \\
\textbf{JNLPBA}                                           & Micro F-1                                    & 93.6 (0.0)                                                                                    & 93.4 (0.0)                                                                                     & \textbf{93.7 (0.1)}                                                                            & \textbf{93.7 (0.0)}                                                                             & 93.6 (0.0)                                                                                      & \textbf{93.7 (0.0)}                                                                     \\
\textbf{}                                                 & Macro F-1                                    & \textbf{79.3 (0.1)}                                                                           & 79.0 (0.1)                                                                                     & 78.5 (0.3)                                                                                     & 78.5 (0.2)                                                                                      & 77.8 (0.1)                                                                                      & 78.2 (0.1)                                                                              \\
\textbf{NCBI-disease}                                     & Micro F-1                                    & 98.3 (0.0)                                                                                    & 98.4 (0.0)                                                                                     & \textbf{98.6 (0.0)}                                                                            & \textbf{98.6 (0.0)}                                                                             & 98.5 (0.0)                                                                                      & \textbf{98.6 (0.0)}                                                                     \\
\textbf{}                                                 & Macro F-1                                    & 93.3 (0.1)                                                                                    & \textbf{93.5 (0.2)}                                                                            & 93.1 (0.1)                                                                                     & 93.3 (0.1)                                                                                      & 92.8 (0.1)                                                                                      & 93.4 (0.1)                                                                              \\ \midrule
\multicolumn{2}{l}{\textbf{\begin{tabular}[c]{@{}l@{}}NER Average \\ Score\end{tabular}}}                & 92.1                                                                                 & 92.2                                                                                           & 91.8                                                                                           & 92.2                                                                                            & 91.9                                                                                            & \textbf{92.3}                                                                           \\ \midrule
\textbf{TriviaQA}                                         & Micro F-1                                    & 78.6 (0.2)                                                                                    & \textbf{79.1 (0.2)}                                                                            & 77.9 (0.2)                                                                                     & 77.4 (0.2)                                                                                      & 77.0 (0.2)                                                                                      & 78.5 (0.1)                                                                              \\ 
\textbf{}                                                 & Macro F-1                                    & 81.6 (0.1)                                                                          & \textbf{82.3 (0.2)}                                                                            & 81.2 (0.1)                                                                                     & 80.6 (0.1)                                                                                      & 80.1 (0.2)                                                                                      & 81.8 (0.1)                                                                              \\ 
\textbf{SQuAD}                                            & Micro F-1                                    & 88.6 (0.0)                                                                                    & 87.2 (0.1)                                                                                     & 88.6 (0.1)                                                                                     & \textbf{88.7 (0.0)}                                                                             & 87.1 (0.0)                                                                                      & 88.5 (0.1)                                                                              \\
                                                          & Macro F-1                                    & \textbf{94.7 (0.0)}                                                                           & 93.9 (0.0)                                                                                     & 94.6 (0.0)                                                                                     & 94.5 (0.0)                                                                                      & 93.5 (0.0)                                                                                      & 94.6 (0.0)                                                                              \\ \midrule
\multicolumn{2}{l}{\textbf{\begin{tabular}[c]{@{}l@{}}QA Average\\ Score\end{tabular}}}                  & \textbf{85.9}                                                                                 & 85.6                                                                                           & 85.6                                                                                           & 85.3                                                                                            & 84.4                                                                                            & 85.8  \\ \bottomrule                                                                                 
\end{tabular}
\caption{DeBERTaV2-XL Results for Reduced Models. \textbf{Bold} indicates the best average score between the standard reduced, adapter-based reduced, and fully fine-tuned versions of each model. \textbf{Reduced + Half Adpt} indicates adapters on the transformer layers of a fully frozen reduced model, where the earlier half of transformer layers were removed and their activations cached. \textbf{Full Adapters} indicates adapters on all transformer layers of a fully frozen model. Each score represents the average score of 5 runs, with the standard errors for each score in parentheses.}
\label{tab:deberta}
\end{table*}

\begin{table*}[htp]
\centering
\small
\setlength{\tabcolsep}{2.3pt}
\begin{tabular}{llrrrrrr}
\toprule
                                                          &                                              & \multicolumn{1}{l}{}                                                                      & \multicolumn{1}{l}{}                                                                       & \multicolumn{4}{c}{\textbf{T5 Large}}                                                                                                                                                                                                                                                                                                                                                        \\ \midrule
                                                          & \textbf{}                                    & \multicolumn{1}{l}{\textbf{\begin{tabular}[c]{@{}l@{}}Reduced +\\ Half Adpt\end{tabular}}} & \multicolumn{1}{l}{\textbf{\begin{tabular}[c]{@{}l@{}} Full \\ Adapters\end{tabular}}} & \multicolumn{1}{l}{\textbf{\begin{tabular}[c]{@{}l@{}}6 Layers \\ Frozen \end{tabular}}} & \multicolumn{1}{l}{\textbf{\begin{tabular}[c]{@{}l@{}}12 Layers \\ Reduced \end{tabular}}} & \multicolumn{1}{l}{\textbf{\begin{tabular}[c]{@{}l@{}}18 Layers \\ Reduced \end{tabular}}} & \multicolumn{1}{l}{\textbf{\begin{tabular}[c]{@{}l@{}}Fully \\ Finetuned\end{tabular}}} \\ \midrule
\textbf{ChemProt}                                         & Micro F-1                                    & 84.3 (0.6)                                                                                & 84.9 (0.6)                                                                                 & 84.7 (0.6)                                                                                     & 84.6 (0.6)                                                                                      & \textbf{85.0 (0.1)}                                                                             & 84.1 (0.8)                                                                              \\
\textbf{}                                                 & Macro F-1                                    & 57.2 (0.7)                                                                                & \textbf{58.0 (0.8)}                                                                        & 56.2 (0.7)                                                                                     & 56.2 (0.7)                                                                                      & 57.4 (0.1)                                                                                      & 56.1 (0.7)                                                                              \\
\textbf{SciCite}                                          & Micro F-1                                    & 86.7 (0.3)                                                                                & 86.2 (0.3)                                                                                 & 87.4 (0.2)                                                                                     & 87.6 (0.1)                                                                                      & \textbf{88.0 (0.2)}                                                                             & 86.4 (0.2)                                                                              \\
\textbf{}                                                 & Macro F-1                                    & 85.3 (0.4)                                                                                & 84.5 (0.4)                                                                                 & 86.0 (0.2)                                                                                     & 86.3 (0.2)                                                                                      & \textbf{86.9 (0.2)}                                                                             & 84.9 (0.2)                                                                              \\
\textbf{SciERC-Rel}                                       & Micro F-1                                    & 85.6 (0.4)                                                                                & 85.2 (0.1)                                                                                 & 84.3 (0.3)                                                                                     & 86.8 (0.4)                                                                                      & 83.4 (0.7)                                                                                      & \textbf{87.4 (0.5)}                                                                     \\
\textbf{}                                                 & Macro F-1                                    & 76.2 (1.0)                                                                                & 75.6 (0.2)                                                                                 & 73.6 (0.9)                                                                                     & 77.4 (0.7)                                                                                      & 72.2 (1.0)                                                                                      & \textbf{80.2 (1.1)}                                                                     \\ \midrule
\multicolumn{2}{l}{\textbf{\begin{tabular}[c]{@{}l@{}}Text Classification\\ Average Score\end{tabular}}} & 79.2                                                                                      & 79.1                                                                                       & 78.7                                                                                           & 79.8                                                                                            & 78.8                                                                                            & \textbf{79.9}                                                                           \\ \midrule
\textbf{bc5cdr}                                           & Micro F-1                                    & 93.8 (0.6)                                                                                & 95.7 (0.7)                                                                                 & \textbf{97.7 (0.7)}                                                                            & 97.4 (0.3)                                                                                      & 95.4 (0.8)                                                                                      & 97.5 (0.2)                                                                              \\
\textbf{}                                                 & Macro F-1                                    & 79.9 (1.0)                                                                                & 85.7 (1.1)                                                                                 & \textbf{91.1 (0.5)}                                                                            & 90.7 (1.1)                                                                                      & 89.3 (1.0)                                                                                      & 89.9 (0.8)                                                                              \\
\textbf{JNLPBA}                                           & Micro F-1                                    & 93.9 (0.4)                                                                                & 93.8 (0.1)                                                                                 & 93.8 (0.0)                                                                                     & 94.0 (0.0)                                                                                      & 93.9 (0.0)                                                                                      & \textbf{94.2 (0.0)}                                                                     \\
\textbf{}                                                 & Macro F-1                                    & 78.8 (0.6)                                                                                & 79.5 (0.2)                                                                                 & 78.8 (0.1)                                                                                     & 79.6 (0.1)                                                                                      & 79.3 (0.0)                                                                                      & \textbf{80.0 (0.0)}                                                                     \\
\textbf{NCBI-disease}                                     & Micro F-1                                    & 97.8 (0.0)                                                                                & 98.5 (0.0)                                                                                 & 98.5 (0.0)                                                                                     & 98.5 (0.0)                                                                                      & 98.4 (0.0)                                                                                      & \textbf{98.6 (0.0)}                                                                     \\
\textbf{}                                                 & Macro F-1                                    & 92.1 (0.2)                                                                                & 92.5 (0.2)                                                                                 & 93.1 (0.1)                                                                                     & 92.8 (0.0)                                                                                      & 92.2 (0.1)                                                                                      & \textbf{93.5 (0.0)}                                                                     \\ \midrule
\multicolumn{2}{l}{\textbf{\begin{tabular}[c]{@{}l@{}}NER Average \\ Score\end{tabular}}}                & 89.4                                                                                      & 90.9                                                                                       & 92.2                                                                                           & 92.2                                                                                            & 91.4                                                                                            & \textbf{92.3}                                                                           \\ \midrule
\textbf{TriviaQA}                                         & Micro F-1                                    & 68.2 (0.2)                                                                                & \textbf{68.8 (0.2)}                                                                        & 67.0 (0.0)                                                                                     & 66.9 (0.0)                                                                                      & 63.9 (0.0)                                                                                      & 68.7 (0.0)                                                                              \\
\textbf{}                                                 & Macro F-1                                    & 77.0 (0.1)                                                                                & 77.5 (0.1)                                                                                 & 77.5 (0.0)                                                                                     & 77.3 (0.0)                                                                                      & 74.8 (0.0)                                                                                      & \textbf{78.0 (0.0)}                                                                     \\
\textbf{SQuAD}                                            & Micro F-1                                    & 81.2 (0.1)                                                                                & 82.0 (0.1)                                                                                 & 86.6 (0.1)                                                                                     & 86.3 (0.6)                                                                                      & 85.2 (0.4)                                                                                      & \textbf{86.7 (0.4)}                                                                     \\
                                                          & Macro F-1                                    & 90.6 (0.1)                                                                                & 91.0 (0.1)                                                                                 & 93.8 (0.0)                                                                                     & 93.7 (0.3)                                                                                      & 92.8 (0.2)                                                                                      & \textbf{93.9 (0.3)}                                                                     \\ \midrule
\multicolumn{2}{l}{\textbf{\begin{tabular}[c]{@{}l@{}}QA Average\\ Score\end{tabular}}}                  & 79.2                                                                                      & 79.8                                                                                       & 81.2                                                                                           & 81.0                                                                                            & 79.2                                                                                            & \textbf{81.8}     \\ \bottomrule                                                                     
\end{tabular}
\caption{T5 Large Results for Reduced Models. \textbf{Bold} indicates the best average score between the standard reduced, adapter-based reduced, and fully fine-tuned versions of each model. \textbf{Reduced + Half Adpt} indicates adapters on the encoder and decoder transformer layers of a fully frozen reduced model, where the earlier half of the encoder layers were removed and their activations cached. \textbf{Full Adapters} indicates adapters on all encoder and decoder transformer layers of a fully frozen model. Each score represents the average score of 5 runs, with the standard errors for each score in parentheses.}
\label{tab:t5_large}
\end{table*}

\begin{table*}[ht]
\centering
\begin{tabular}{llrrrr}
\toprule
                                                       &                                                       & \multicolumn{4}{c}{\textbf{DistilBERT}}                                                                                                                                                                                                                                                                                                                                                        \\ \midrule
                                                       & \textbf{}                                  & \multicolumn{1}{l}{\textbf{\begin{tabular}[c]{@{}l@{}}2 Layers \\ Reduced\end{tabular}}} & \multicolumn{1}{l}{\textbf{\begin{tabular}[c]{@{}l@{}}3 Layers \\ Reduced\end{tabular}}} & \multicolumn{1}{l}{\textbf{\begin{tabular}[c]{@{}l@{}}4 Layers \\ Reduced\end{tabular}}} & \multicolumn{1}{l}{\textbf{\begin{tabular}[c]{@{}l@{}}Fully \\ Fine-tuned\end{tabular}}} \\ \midrule
\textbf{ChemProt}                                      & Micro F-1                                             & 79.1 (0.4)                                                                                       & \textbf{80.3 (0.1)}                                                                            & 79.0 (0.2)                                                                                       & 79.1 (0.5)                                                                              \\
\textbf{}                                              & Macro F-1                                             & 52.1 (0.5)                                                                                       & 51.6 (0.6)                                                                                     & 51.6 (0.4)                                                                                       & \textbf{52.6 (0.3)}                                                                     \\
\textbf{SciCite}                                       & Micro F-1                                             & 85.7 (0.1)                                                                                       & 85.6 (0.1)                                                                                     & \textbf{85.8 (0.1)}                                                                              & 85.5 (0.1)                                                                              \\
\textbf{}                                              & Macro F-1                                             & \textbf{84.3 (0.1)}                                                                              & 84.1 (0.1)                                                                                     & 84.2 (0.1)                                                                                       & 84.0 (0.1)                                                                              \\
\textbf{SciERC-Rel}                                    & Micro F-1                                             & 84.3 (0.3)                                                                                       & 84.5 (0.3)                                                                                     & \textbf{84.6 (0.2)}                                                                              & 83.5 (0.4)                                                                              \\
\textbf{}                                              & Macro F-1                                             & 74.1 (0.7)                                                                                       & \textbf{74.9 (0.7)}                                                                            & 74.6 (0.4)                                                                                       & 72.9 (0.7)                                                                              \\ \midrule
\multicolumn{2}{l}{\textbf{\begin{tabular}[c]{@{}l@{}}Text Classification\\ Average Score\end{tabular}}} & 76.6                                                                                             & \textbf{76.8}                                                                                  & 76.6                                                                                             & 76.3                                                                                    \\ \midrule
\textbf{bc5cdr}                                        & Micro F-1                                             & 97.0 (0.0)                                                                                       & 97.0 (0.0)                                                                                     & 96.9 (0.0)                                                                                       & \textbf{97.2 (0.0)}                                                                     \\
\textbf{}                                              & Macro F-1                                             & 88.3 (0.0)                                                                                       & 88.3 (0.1)                                                                                     & 87.9 (0.0)                                                                                       & \textbf{88.7 (0.1)}                                                                     \\
\textbf{JNLPBA}                                        & Micro F-1                                             & 93.4 (0.1)                                                                                       & \textbf{93.5 (0.0)}                                                                            & 93.4 (0.0)                                                                                       & \textbf{93.5 (0.0)}                                                                     \\
\textbf{}                                              & Macro F-1                                             & 78.0 (0.3)                                                                                       & \textbf{78.6 (0.1)}                                                                            & 77.9 (0.1)                                                                                       & 78.5 (0.1)                                                                              \\
\textbf{NCBI-disease}                                  & Micro F-1                                             & \textbf{98.2 (0.0)}                                                                              & 98.0 (0.0)                                                                                     & 98.1 (0.0)                                                                                       & \textbf{98.2 (0.0)}                                                                     \\
\textbf{}                                              & Macro F-1                                             & \textbf{91.4 (0.1)}                                                                              & 90.5 (0.1)                                                                                     & 90.7 (0.1)                                                                                       & 91.3 (0.1)                                                                              \\ \midrule
\multicolumn{2}{l}{\textbf{\begin{tabular}[c]{@{}l@{}}NER Average \\ Score\end{tabular}}}                & 91.1                                                                                             & 91                                                                                             & 90.8                                                                                             & \textbf{91.2}                                                                           \\ \midrule
\textbf{TriviaQA}                                      & Micro F-1                                             & 62.9 (0.1)                                                                                       & 61.4 (0.1)                                                                                     & 59.1 (0.1)                                                                                       & \textbf{63.6 (0.1)}                                                                     \\
                                                       & Macro F-1                                             & 66.2 (0.1)                                                                                       & 64.7 (0.1)                                                                                     & 62.4 (0.1)                                                                                       & \textbf{66.8 (0.1)}                                                                     \\
\textbf{SQuAD}                                         & Micro F-1                                             & 76.6 (0.1)                                                                                       & 76.3 (0.1)                                                                                     & 72.5 (0.1)                                                                                       & \textbf{77.1 (0.1)}                                                                     \\
                                                       & Macro F-1                                             & 85.1 (0.1)                                                                                       & 84.8 (0.0)                                                                                     & 82.3 (0.1)                                                                                       & \textbf{85.4 (0.0)}                                                                     \\ \midrule
\multicolumn{2}{l}{\textbf{\begin{tabular}[c]{@{}l@{}}QA Average\\ Score\end{tabular}}}                  & 72.7                                                                                             & 71.8                                                                                           & 69.1                                                                                             & \textbf{73.2} \\ \bottomrule                                                                         
\end{tabular}
\caption{DistilBERT Results for Reduced Models. \textbf{Bold} indicates the best average score between the reduced and fully fine-tuned versions of each model. Each score represents the average score of 10 runs, with the standard errors for each score in parentheses.}
\label{tab:distilbert}
\end{table*}

\begin{table*}[ht]
\centering
\begin{tabular}{llrrrr}
\toprule
                                                       &                                                       & \multicolumn{4}{c}{\textbf{MiniLM: 6L-H768}}                                                                                                                                                                                                                                                                                                                                                   \\ \midrule
                                                       & \textbf{}                                  & \multicolumn{1}{l}{\textbf{\begin{tabular}[c]{@{}l@{}}2 Layers \\ Reduced\end{tabular}}} & \multicolumn{1}{l}{\textbf{\begin{tabular}[c]{@{}l@{}}3 Layers \\ Reduced\end{tabular}}} & \multicolumn{1}{l}{\textbf{\begin{tabular}[c]{@{}l@{}}4 Layers \\ Reduced\end{tabular}}} & \multicolumn{1}{l}{\textbf{\begin{tabular}[c]{@{}l@{}}Fully \\ Fine-tuned\end{tabular}}} \\ \midrule
\textbf{ChemProt}                                      & Micro F-1                                             & \textbf{79.4 (0.3)}                                                                              & 78.3 (0.4)                                                                                     & 79.0 (0.2)                                                                                       & 79.3 (0.3)                                                                              \\
\textbf{}                                              & Macro F-1                                             & 51.8 (0.4)                                                                                       & 50.6 (0.4)                                                                                     & 52.0 (0.2)                                                                                       & \textbf{52.6 (0.4)}                                                                     \\
\textbf{SciCite}                                       & Micro F-1                                             & 85.4 (0.1)                                                                                       & 85.8 (0.2)                                                                                     & 85.9 (0.1)                                                                                       & \textbf{86.0 (0.2)}                                                                     \\
\textbf{}                                              & Macro F-1                                             & 84.1 (0.2)                                                                                       & 84.5 (0.2)                                                                                     & 84.5 (0.1)                                                                                       & \textbf{84.6 (0.2)}                                                                     \\
\textbf{SciERC-Rel}                                    & Micro F-1                                             & 84.7 (0.3)                                                                                       & 83.9 (0.3)                                                                                     & 84.1 (0.4)                                                                                       & \textbf{86.3 (0.2)}                                                                     \\
\textbf{}                                              & Macro F-1                                             & 75.0 (0.4)                                                                                       & 74.8 (0.4)                                                                                     & 75.3 (0.6)                                                                                       & \textbf{78.2 (0.6)}                                                                     \\ \midrule
\multicolumn{2}{l}{\textbf{\begin{tabular}[c]{@{}l@{}}Text Classification\\ Average Score\end{tabular}}} & 76.7                                                                                             & 76.3                                                                                           & 76.8                                                                                             & \textbf{77.8}                                                                           \\ \midrule
\textbf{bc5cdr}                                        & Micro F-1                                             & 96.1 (0.3)                                                                                       & \textbf{96.8 (0.0)}                                                                            & 96.6 (0.0)                                                                                       & \textbf{96.8 (0.2)}                                                                     \\
\textbf{}                                              & Macro F-1                                             & 84.6 (1.1)                                                                                       & \textbf{87.8 (0.1)}                                                                            & 86.6 (0.0)                                                                                       & 87.5 (1.0)                                                                              \\
\textbf{JNLPBA}                                        & Micro F-1                                             & 93.2 (0.0)                                                                                       & 93.2 (0.0)                                                                                     & \textbf{93.3 (0.0)}                                                                              & \textbf{93.3 (0.0)}                                                                     \\
\textbf{}                                              & Macro F-1                                             & \textbf{77.5 (0.1)}                                                                              & 77.3 (0.1)                                                                                     & 77.3 (0.1)                                                                                       & 76.9 (0.2)                                                                              \\
\textbf{NCBI-disease}                                  & Micro F-1                                             & \textbf{98.3 (0.0)}                                                                              & 98.2 (0.0)                                                                                     & 98.2 (0.0)                                                                                       & \textbf{98.3 (0.0)}                                                                     \\
\textbf{}                                              & Macro F-1                                             & \textbf{92.1 (0.1)}                                                                              & 91.1 (0.1)                                                                                     & 91.0 (0.1)                                                                                       & \textbf{92.1 (0.1)}                                                                     \\ \midrule
\multicolumn{2}{l}{\textbf{\begin{tabular}[c]{@{}l@{}}NER Average \\ Score \end{tabular}}}                & 90.3                                                                                             & 90.7                                                                                           & 90.5                                                                                             & \textbf{90.8}                                                                           \\ \midrule
\textbf{TriviaQA}                                      & Micro F-1                                             & 70.2 (0.1)                                                                                       & 68.9 (0.1)                                                                                     & 65.5 (0.1)                                                                                       & \textbf{70.4 (0.2)}                                                                     \\
                                                       & Macro F-1                                             & 73.4 (0.1)                                                                                       & 72.2 (0.1)                                                                                     & 68.9 (0.1)                                                                                       & \textbf{73.8 (0.2)}                                                                     \\
\textbf{SQuAD}                                         & Micro F-1                                             & 77.6 (0.1)                                                                                       & 75.6 (0.1)                                                                                     & 65.4 (0.2)                                                                                       & \textbf{78.9 (0.1)}                                                                     \\
                                                       & Macro F-1                                             & 86.4 (0.1)                                                                                       & 85.0 (0.1)                                                                                     & 77.0 (0.1)                                                                                       & \textbf{87.0 (0.1)}                                                                     \\ \midrule
\multicolumn{2}{l}{\textbf{\begin{tabular}[c]{@{}l@{}}QA Average\\ Score\end{tabular}}}                  & 76.9                                                                                             & 75.4                                                                                           & 69.2                                                                                             & \textbf{77.5}  \\ \bottomrule                                                                        
\end{tabular}
\caption{MiniLM L6-H768 Results for Reduced Models. \textbf{Bold} indicates the best average score between the reduced and fully fine-tuned versions of each model. Each score represents the average score of 10 runs, with the standard errors for each score in parentheses.}
\label{tab:minilm_768}
\end{table*}

\begin{table*}[ht]
\centering
\begin{tabular}{llrrrr}
\toprule
                                                       &                                                       & \multicolumn{4}{c}{\textbf{MiniLM: L6-H384}}                                                                                                                                                                                                                                                                                                                                                   \\ \midrule
                                                       & \textbf{}                                  & \multicolumn{1}{l}{\textbf{\begin{tabular}[c]{@{}l@{}}2 Layers \\ Reduced\end{tabular}}} & \multicolumn{1}{l}{\textbf{\begin{tabular}[c]{@{}l@{}}3 Layers \\ Reduced\end{tabular}}} & \multicolumn{1}{l}{\textbf{\begin{tabular}[c]{@{}l@{}}4 Layers \\ Reduced\end{tabular}}} & \multicolumn{1}{l}{\textbf{\begin{tabular}[c]{@{}l@{}}Fully \\ Fine-tuned\end{tabular}}} \\ \midrule
\textbf{ChemProt}                                      & Micro F-1                                             & 75.4 (0.5)                                                                                       & \textbf{76.9 (0.2)}                                                                            & 74.9 (0.3)                                                                                       & 74.6 (0.4)                                                                              \\
\textbf{}                                              & Macro F-1                                             & 47.3 (0.7)                                                                                       & \textbf{50.4 (0.2)}                                                                            & 48.8 (0.4)                                                                                       & 47.1 (0.8)                                                                              \\
\textbf{SciCite}                                       & Micro F-1                                             & 84.4 (0.1)                                                                                       & \textbf{85.4 (0.1)}                                                                            & 85.1 (0.1)                                                                                       & 84.4 (0.1)                                                                              \\
\textbf{}                                              & Macro F-1                                             & 82.8 (0.1)                                                                                       & \textbf{83.7 (0.1)}                                                                            & 83.4 (0.1)                                                                                       & 82.8 (0.1)                                                                              \\
\textbf{SciERC-Rel}                                    & Micro F-1                                             & 83.2 (0.3)                                                                                       & 82.6 (0.3)                                                                                     & \textbf{83.3 (0.2)}                                                                              & 79.5 (0.9)                                                                              \\
\textbf{}                                              & Macro F-1                                             & 72.7 (0.6)                                                                                       & 72.1 (0.6)                                                                                     & \textbf{73.7 (0.3)}                                                                              & 68.9 (1.1)                                                                              \\ \midrule
\multicolumn{2}{l}{\textbf{\begin{tabular}[c]{@{}l@{}}Text Classification\\ Average Score\end{tabular}}} & 74.3                                                                                             & \textbf{75.2}                                                                                  & 74.9                                                                                             & 72.9                                                                                    \\ \midrule
\textbf{bc5cdr}                                        & Micro F-1                                             & 96.6 (0.0)                                                                                       & 96.3 (0.0)                                                                                     & 95.6 (0.0)                                                                                       & \textbf{96.9 (0.0)}                                                                     \\
\textbf{}                                              & Macro F-1                                             & 86.9 (0.1)                                                                                       & 85.9 (0.1)                                                                                     & 83.2 (0.1)                                                                                       & \textbf{88.3 (0.1)}                                                                     \\
\textbf{JNLPBA}                                        & Micro F-1                                             & 93.0 (0.0)                                                                                       & 92.2 (0.0)                                                                                     & 92.0 (0.0)                                                                                       & \textbf{93.3 (0.0)}                                                                     \\
\textbf{}                                              & Macro F-1                                             & 76.3 (0.1)                                                                                       & 74.0 (0.1)                                                                                     & 73.6 (0.1)                                                                                       & \textbf{77.2 (0.1)}                                                                     \\
\textbf{NCBI-disease}                                  & Micro F-1                                             & 98.0 (0.0)                                                                                       & 97.9 (0.0)                                                                                     & 97.7 (0.0)                                                                                       & \textbf{98.2 (0.0)}                                                                     \\
\textbf{}                                              & Macro F-1                                             & 90.6 (0.1)                                                                                       & 89.9 (0.1)                                                                                     & 88.9 (0.1)                                                                                       & \textbf{91.7 (0.1)}                                                                     \\ \midrule
\multicolumn{2}{l}{\textbf{\begin{tabular}[c]{@{}l@{}}NER Average \\ Score\end{tabular}}}                & 90.2                                                                                             & 89.4                                                                                           & 88.5                                                                                             & \textbf{90.9}                                                                           \\ \midrule
\textbf{TriviaQA}                                      & Micro F-1                                             & 66.6 (0.1)                                                                                       & 65.6 (0.1)                                                                                     & 63.4 (0.1)                                                                                       & \textbf{67.6 (0.2)}                                                                     \\
\textbf{}                                              & Macro F-1                                             & 69.9 (0.1)                                                                                       & 69.2 (0.1)                                                                                     & 67.0 (0.1)                                                                                       & \textbf{71.0 (0.2)}                                                                     \\
\textbf{SQuAD}                                         & Micro F-1                                             & \textbf{81.6 (0.0)}                                                                              & 80.9 (0.1)                                                                                     & 74.2 (0.2)                                                                                       & \textbf{81.6 (0.1)}                                                                     \\
                                                       & Macro F-1                                             & \textbf{89.7 (0.0)}                                                                              & 89.0 (0.0)                                                                                     & 84.5 (0.1)                                                                                       & 89.6 (0.0)                                                                              \\ \midrule
\multicolumn{2}{l}{\textbf{\begin{tabular}[c]{@{}l@{}}QA Average\\ Score \end{tabular}}}                  & 76.9                                                                                             & 76.2                                                                                           & 72.3                                                                                             & \textbf{77.4}    \\ \bottomrule                                                                      
\end{tabular}
\caption{MiniLM L6-H384 Results for Reduced Models. \textbf{Bold} indicates the best average score between the reduced and fully fine-tuned versions of each model. Each score represents the average score of 10 runs, with the standard errors for each score in parentheses.}
\label{tab:minilm_384}
\end{table*}

\begin{table*}[!htp]
\footnotesize
\centering
\normalsize
\begin{tabular}{lrrr}
\toprule
\multicolumn{1}{c}{\textbf{Task}} & \multicolumn{1}{c}{\textbf{Averages}} & \multicolumn{1}{c}{\textbf{\begin{tabular}[c]{@{}l@{}} Standard \\ Recycling\end{tabular}}} & \textbf{ \begin{tabular}[c]{@{}l@{}} Adapter-Based \\ Recycling\end{tabular}} \\
\midrule
\textbf{Classification} & \begin{tabular}[c]{@{}l@{}} Training Time\end{tabular} & 2204  & 2349 \\
 & \begin{tabular}[c]{@{}l@{}}Epochs\end{tabular} & 38 & 42 \\
\midrule
\textbf{NER} & \begin{tabular}[c]{@{}l@{}} Training Time\end{tabular} & 4269 & 3857  \\
 & \begin{tabular}[c]{@{}l@{}}Epochs\end{tabular} & 43 & 39 \\
\midrule
\textbf{QA} & \begin{tabular}[c]{@{}l@{}} Training Time\end{tabular} & 8252 & 8513 \\
 & \begin{tabular}[c]{@{}l@{}}Epochs\end{tabular} & 6 & 7 \\
\bottomrule
\end{tabular}
\caption{Average Training Times and Epochs for Embedding Recycling (seconds for training time, count for epochs). \textbf{Standard Recycling} corresponds to layer recycling on a reduced transformer model. \textbf{Adapter-Based Recycling} corresponds to layer recycling on a reduced frozen transformer model with added trainable Adapter modules. Training time and epoch averages are the averages across the RoBERTa, BERT, SciBERT, DeBERTa V2 XL, and T5-Large transformer models and the text classification, NER, and QA datasets tested.}
\label{tab:training_times}
\end{table*}

\end{document}